\documentclass[11pt]{article}

\usepackage[a4paper,margin=1in]{geometry}
\usepackage{graphicx}
\usepackage{amsmath,amssymb}
\usepackage{booktabs}
\usepackage{array}
\usepackage{multirow}
\usepackage{xcolor}
\usepackage{colortbl}
\usepackage[pagebackref,breaklinks,colorlinks,allcolors=cvprblue]{hyperref}
\usepackage[most]{tcolorbox}
\tcbset{
  breakable,
  colback=gray!5,
  boxrule=0.4pt,
  arc=0pt,
  outer arc=0pt,
}
\definecolor{cvprblue}{rgb}{0.21,0.49,0.74}

\usepackage{CJKutf8}
\newcommand{\zhtext}[2][-0.15ex]{%
  \raisebox{#1}{\begin{CJK*}{UTF8}{gkai}#2\end{CJK*}}%
}

\usepackage{listings}
\usepackage{inconsolata}
\usepackage{makecell}

\lstdefinestyle{promptstyle}{
    basicstyle=\ttfamily\small, % inconsolata
    breaklines=true,
    frame=single,
    backgroundcolor=\color{gray!5},
    postbreak=\mbox{\textcolor{cvprblue}{$\hookrightarrow$}\space},
    columns=fullflexible,
}
\lstdefinestyle{casestyle}{
    basicstyle=\ttfamily\small, % inconsolata
    breaklines=true,
    frame=single,
    backgroundcolor=\color{gray!20},
    postbreak=\mbox{\textcolor{cvprblue}{$\hookrightarrow$}\space},
    columns=fullflexible,
}

\title{\raisebox{-0.6em}{\includegraphics[height=2em]{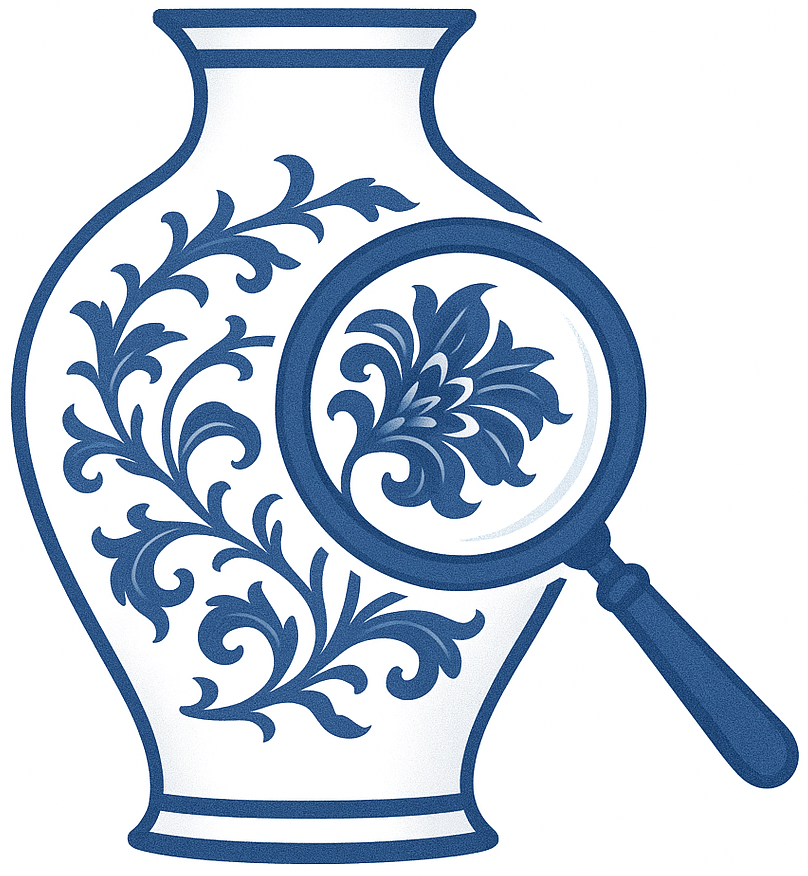}} \,  CiQi-Agent: Aligning Vision, Tools and Aesthetics in Multimodal Agent for Cultural Reasoning on Chinese Porcelains}

\author{
Wenhan Wang\thanks{Equal contribution.} $^{,1}$ \quad Zhixiang Zhou\footnotemark[1] $^{,1}$ \quad Zhongtian Ma\footnotemark[1] $^{,2}$ \quad Yanzhu Chen$^{1}$ \\ Ziyu Lin$^{1}$ \quad Hao Sheng$^{1}$\quad Pengfei Liu$^{1}$ \quad Honglin Ma$^{3}$ \quad Wenqi Shao\thanks{Corresponding to: Wenqi Shao (\href{mailto:weqish@gmail.com}{weqish@gmail.com}), Qiaosheng Zhang (\href{mailto:zhangqiaosheng@pjlab.org.cn}{zhangqiaosheng@pjlab.org.cn}).} $^{,1}$ \\ Qiaosheng Zhang\footnotemark[2] $^{,1,2}$ \quad Yu Qiao$^{1,2}$
\\[0.8em]
$^{1}$ Shanghai Innovation Institute \quad
$^{2}$ Shanghai AI Laboratory \\
$^{3}$ Shaanxi Academy of Cultural Relics Conservation
}

\date{} % 不显示日期

\begin{document}
\maketitle

\begin{abstract}
The connoisseurship of antique Chinese porcelain demands extensive historical expertise, material understanding, and aesthetic sensitivity, making it difficult for non-specialists to engage. To democratize cultural-heritage understanding and assist expert connoisseurship, we introduce \textbf{CiQi-Agent}---a domain-specific \textbf{Porcelain Connoisseurship Agent} for intelligent analysis of antique Chinese porcelain. 
CiQi-Agent supports multi-image porcelain inputs and enables vision tool invocation and multimodal retrieval-augmented generation, performing fine-grained connoisseurship analysis across \textbf{six attributes}: dynasty, reign period, kiln site, glaze color, decorative motif, and vessel shape. Beyond attribute classification, it captures subtle visual details, retrieves relevant domain knowledge, and integrates visual and textual evidence to produce coherent, explainable connoisseurship descriptions.
To achieve this capability, we construct a large-scale, expert-annotated dataset \textbf{CiQi-VQA}, comprising 29,596 porcelain specimens, 51,553 images, and 557,940 visual question–answering pairs, and further establish a comprehensive benchmark \textbf{CiQi-Bench} aligned with the previously mentioned six attributes.
CiQi-Agent is trained through supervised fine-tuning, reinforcement learning, and a tool-augmented reasoning framework that integrates two categories of tools: a vision tool and multimodal retrieval tools.
% {\color{red}
Experimental results show that CiQi-Agent (7B) outperforms all competitive open- and closed-source models across all six attributes on CiQi-Bench, achieving on average \textbf{12.2\%} higher accuracy than GPT-5.
% } 
The model and dataset have been released and are publicly available at \url{https://huggingface.co/datasets/SII-Monument-Valley/CiQi-VQA}.

\end{abstract}

\section{Introduction}
\begin{figure*}[t]
  \centering
   \includegraphics[width=\linewidth]{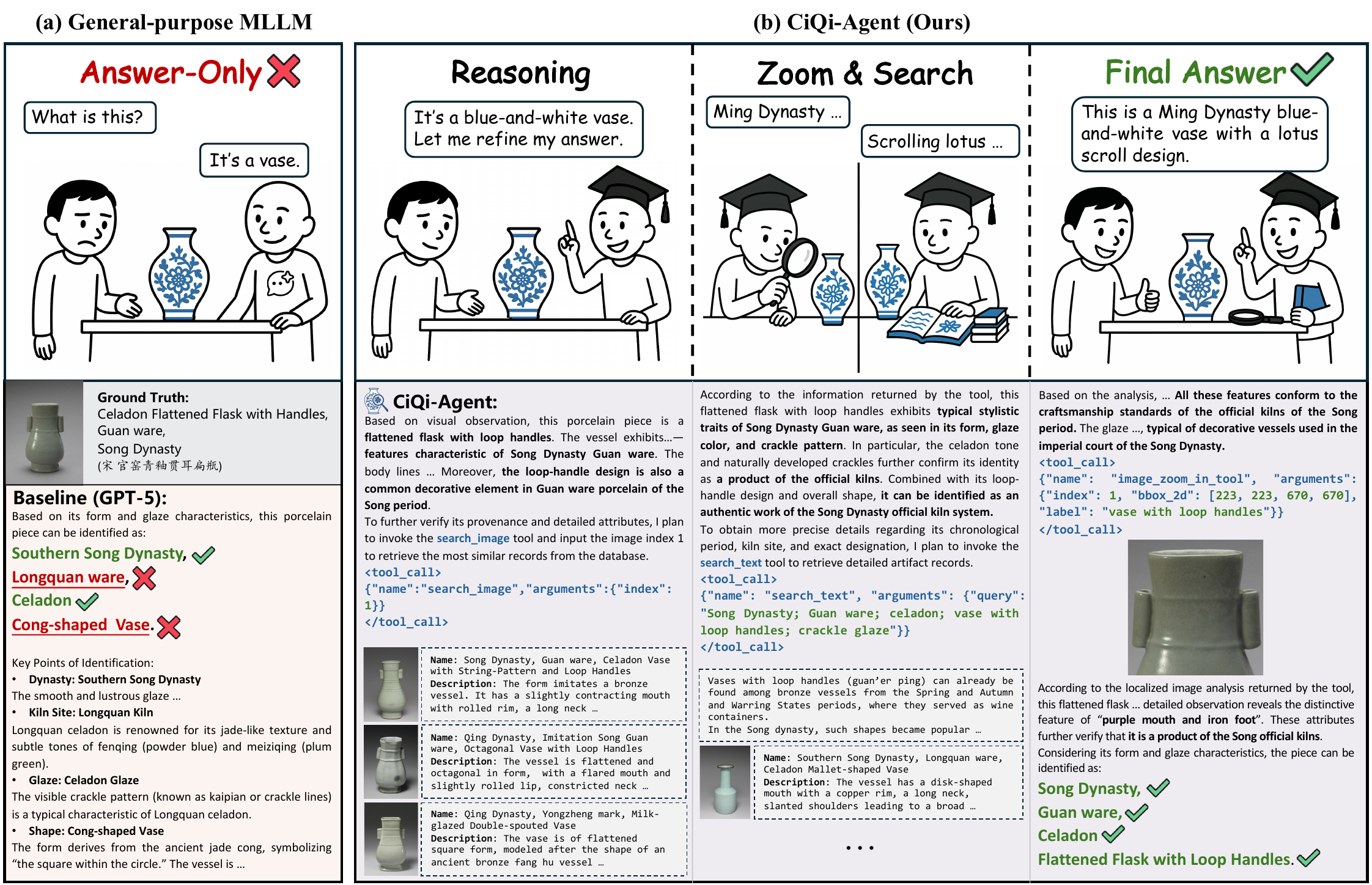}
   \caption{
        \textbf{Comparison between (a) General-purpose MLLM and (b) CiQi-Agent.}
        (a) Conventional MLLMs rely on \textit{single-pass answering}, directly outputting a label leading to superficial or inaccurate identifications. 
        (b) The proposed \textbf{CiQi-Agent} introduces \textit{Tool-Augmented Reasoning}, enabling multi-step porcelain analysis through image zoom-in, image/text retrieval.
        This iterative process yields more reliable final answers aligned with expert connoisseurship reasoning.
        }
   \label{fig:main_fig}
\end{figure*}

Cultural relics are invaluable carriers of human civilization, encapsulating both artistic creation and historical evolution. The connoisseurship and authentication of artifacts require deep expertise—combining historical knowledge, perceptual experience, and material understanding. Consequently, professional barriers have long restricted public engagement in cultural heritage connoisseurship. With the rapid progress of artificial intelligence (AI) methods, especially large language models (LLMs) and multimodal large language models (MLLMs)~\cite{Li2023BLIP2, Liu2023LLaVA, wang2024qwen2vlenhancingvisionlanguagemodels, internvl}, new opportunities have emerged to democratize artifact understanding. By jointly modeling vision, language, and reasoning, MLLMs enable interactive, explainable, and scalable analysis of visual art and historical objects~\cite{10.1145/3664647.3680649, mohamed-etal-2024-culture, Balauca_2025_ICCV, zhou-etal-2025-hanfu}.
In this work, we focus on \textbf{Antique Chinese Porcelain}, one of the most representative yet technically challenging categories of cultural relics, as our entry point for AI-driven artifact connoisseurship.

Existing approaches to porcelain connoisseurship remain limited in several key aspects. In most computer vision (CV) studies, porcelain connoisseurship is simplified as a fine-grained recognition task~\cite{Liu2022porcelain,liu2022where, Ling2025MultiTask, hu2025integrating}, focusing mainly on visual classification without incorporating language-based reasoning or interactive explanation. Moreover, general-purpose MLLMs perform poorly in this specialized domain due to the scarcity of porcelain-related data, leading to weak generalization, unstable judgments, and the absence of a unified evaluation standard, resulting in outputs that are often incomplete and lacking in professionalism. These limitations highlight the need for a dedicated \textit{porcelain connoisseurship agent} that integrates perception, reasoning, and cultural interpretation.

Building such an agent presents several unique challenges.
First, the number of valuable porcelain artifacts is inherently limited, making high-quality data collection extremely difficult. Second, accurate annotation and description of porcelain pieces require deep domain expertise in art history and craftsmanship, which makes large-scale labeling both costly and inconsistent. Third, there is no unified evaluation standard for porcelain connoisseurship, leaving MLLMs without reliable metrics to assess their performance. Finally, the essence of porcelain connoisseurship lies in \textit{fine-grained recognition}, which demands precise identification of features such as vessel shape, glaze color, decorative motifs, and historical period---a task that remains highly challenging even for human experts.

% 在这篇文章中，我们构建了第一个porcelain connoisseurship agent，名为CiQi-Agent。为了解决上述难题，CiQi-Agent构建的核心包括：1.构建瓷器领域数据集与benchmark，2.设计two phase的包含SFT和RL的迭代训练方法，3.加入视觉工具调用与多模态RAG辅助判断。CiQi-Agent的能力包括支持多图输入；支持image-zoom-in、图像检索与文字检索三种工具调用；结合工具调用，给出瓷器在六个属性（xx,xx...）上的鉴赏结果，并且附加专业解释。

To overcome these challenges, we propose \textbf{CiQi-Agent}, the first Chinese porcelain connoisseurship agent that integrates domain-grounded data curation, a two-phase training paradigm combining supervised fine-tuning (SFT) and reinforcement learning (RL), and tool-augmented reasoning into a unified framework (as shown in \ref{fig:main_fig}). CiQi-Agent supports multi-image input and performs fine-grained connoisseurship over six attributes (dynasty, reign period, kiln site, glaze color, decorative motif and vessel shape) by natively integrating vision tool invocation and multimodal retrieval-augmented generation (RAG) as essential reasoning capabilities, and can optionally output the top-$k$ visually similar porcelains to facilitate human interpretation and reference. 
Specifically, our contributions are as follows:

% To address these challenges, we propose CiQi-Agent, a domain-specialized multimodal agent for Chinese porcelain connoisseurship. CiQi-Agnet is trained with supervised fine-tuning (SFT) and reinforcement learning (RL), augmented by visual tool invocation and retrieval-augmented generation (RAG). 
% We also construct a large-scale expert-annotated porcelain dataset and establish a comprehensive benchmark for porcelain connoisseurship. 
% Experimental results demonstrate that our model substantially outperforms Mainstream open- and closed-source MLLMs on this benchmark. More specifically, our contributions can be summarized as follows:

% \begin{figure*}[t]
%   \centering
%    \includegraphics[width=1\linewidth]{sec/figures/porcelain_distribution.png}
%    \caption{
%         \textbf{Visualization of the four key attributes in porcelain connoisseurship.}
%         Shown are the distributions of dynasty, glaze color, vessel shape, and decorative motif in the raw porcelain dataset. 
%         For each attribute, the top 10 most frequent categories are presented.
%         }
%    \label{fig:dataset_distribution}
% \end{figure*}

\begin{itemize}
   \item We construct \textbf{CiQi-VQA}, a large-scale dataset for ancient Chinese porcelain connoisseurship, with $29{,}596$ original artifacts from $20$+ dynasties (2nd c. BCE–19th c. CE), covering $100$+ vessel shapes, $200$+ glaze colors, and $200$+ decorative motifs. To support multimodal model training, we further expand it to over $500$K high-quality visual question answering (VQA) pairs via a hybrid pipeline combining expert annotation and LLM-assisted cleaning.

    \item We establish \textbf{CiQi-Bench}, an expert-aligned benchmark for Chinese porcelain connoisseurship, comprising $775$ high-quality specimens and two complementary evaluation protocols: (1) a fine-grained multiple-choice setting covering six attributes and standardized naming, and (2) a free-form generation setting assessed via LLM-based attribute-wise similarity scoring.

    % \item We develop a training pipeline that integrates supervised fine-tuning (SFT), reinforcement learning (RL), and tool-augmented reasoning, incorporating visual tool invocation and retrieval-augmented generation (RAG). This framework enables multi-image input and tool-assisted judgment for fine-grained porcelain connoisseurship.
    \item We propose a two-phase iterative training framework for CiQi-Agent: Phase I uses GRPO-based RL with a large tool-calling reward to rapidly bootstrap tool-calling skills and generate synthetic trajectories; Phase II integrates these trajectories back into SFT, followed by RL with a reweighted, accuracy-conditioned reward that jointly optimizes tool-calling proficiency and connoisseurship accuracy, thereby aligning CiQi-Agent’s tool-calling capability with its domain expertise.

    \item CiQi-Agent incorporates both visual and retrieval-based tools, including an image zoom-in tool and image/text retrieval tools to enable multimodal RAG. The retrieval database is primarily built from the CiQi-VQA dataset, consisting of $8{,}161$ curated porcelain pieces with $16{,}380$ images, supplemented by $49{,}606$ cleaned plain-text entries from professional articles.

    \item The proposed CiQi-Agent, built upon the Qwen2.5-VL-7B-Instruct, achieves superior performance on our benchmark, consistently outperforming all mainstream open- and closed-source multimodal models across all evaluation attributes.
\end{itemize}

\section{Related Work}

\medskip
\noindent\textbf{Classical AI Methods for Porcelain Connoisseurship.}
Early attempts at automating porcelain connoisseurship treated it as a fine-grained image classification problem. Traditional CV methods relied on hand-crafted features (e.g., color, texture descriptors) combined with classifier such as support vector machine (SVM), achieving moderate success on small datasets~\cite{Weng2017MachineVision,pavithra2021texture}. With the rise of deep learning, convolutional neural network (CNN)-based approaches have shown improved accuracy by automatically learning visual features like glaze color and vessel shape~\cite{Liu2022porcelain,liu2022where}. Researchers have also explored multi‐task models to jointly classify attributes such as dynasty, kiln, and glaze type. For instance, Ling \emph{et al.} present a deep-learning framework for four attributes (dynasty, glaze, ware, type) on Song/Yuan porcelain~\cite{Ling2025MultiTask}. Nonetheless, these classification-driven works are limited to predefined labels and cannot perform the richer, explanatory reasoning that true connoisseurship demands.  They also typically treat artifact analysis as one-shot classification, rather than interactive or multi-step reasoning~\cite{Mitra_2024_CVPR,chen2024visual,Dong_2025_CVPR}. In summary, classical AI methods underline the need for larger high-quality datasets and for moving beyond one-shot classification toward interactive, knowledge-rich analysis.

\medskip
\noindent\textbf{MLLMs and Domain-Specific Multimodal Systems.}
General-purpose multimodal LLMs such as BLIP-2~\cite{Li2023BLIP2} and LLaVA~\cite{Liu2023LLaVA} bridge image and text understanding, but typically operate in a single-pass manner without iterative perception or external-knowledge integration~\cite{Suris_2023_ICCV,Gupta_2023_CVPR,hu2024visual}. Recent visual-agent systems address these limitations by enabling multi-step perception and tool-based reasoning~\cite{Wu2023VisualChatGPT,Zheng2025DeepEyes,Qiao2025VThinker,Chern2025TWGI}.  
In parallel, domain-adapted multimodal models have emerged in medicine~\cite{Nisar2024DRax,Moor2023ML4H,Tu2024NEJMAI}, remote sensing~\cite{Weng2025RemoteVLM,Jiang2025ICCAD}, and cultural heritage analysis such as VaseVQA~\cite{Ge2025VaseVQA,11179498,Ma2025ECCV}. However, these systems primarily perform \emph{single-step prediction} and do not support tool-augmented reasoning. This gap underscores the need for domain-specific visual agents capable of multi-step perception and knowledge-grounded decision making.

\section{Dataset and Benchmark}
% 本章我们首先介绍我们收集到的原型瓷器数据，以及其在年代、釉色、纹饰内容与器型几个维度上的统计信息。之后我们介绍了如何使用专家标注与LLM辅助的方式构建开放式的VQA的数据集。最后我们介绍了我们提出的benchmark。
% 数据部分：1. 数据的种类与分布。（多少个原型瓷器，怎么分布的，多少张图）2. 用作训练和测试的数据划分。3. 如何清洗的（给出例子和示意图），有image-caption有questions
\begin{figure*}[t]
  \centering
   \includegraphics[width=1\linewidth]{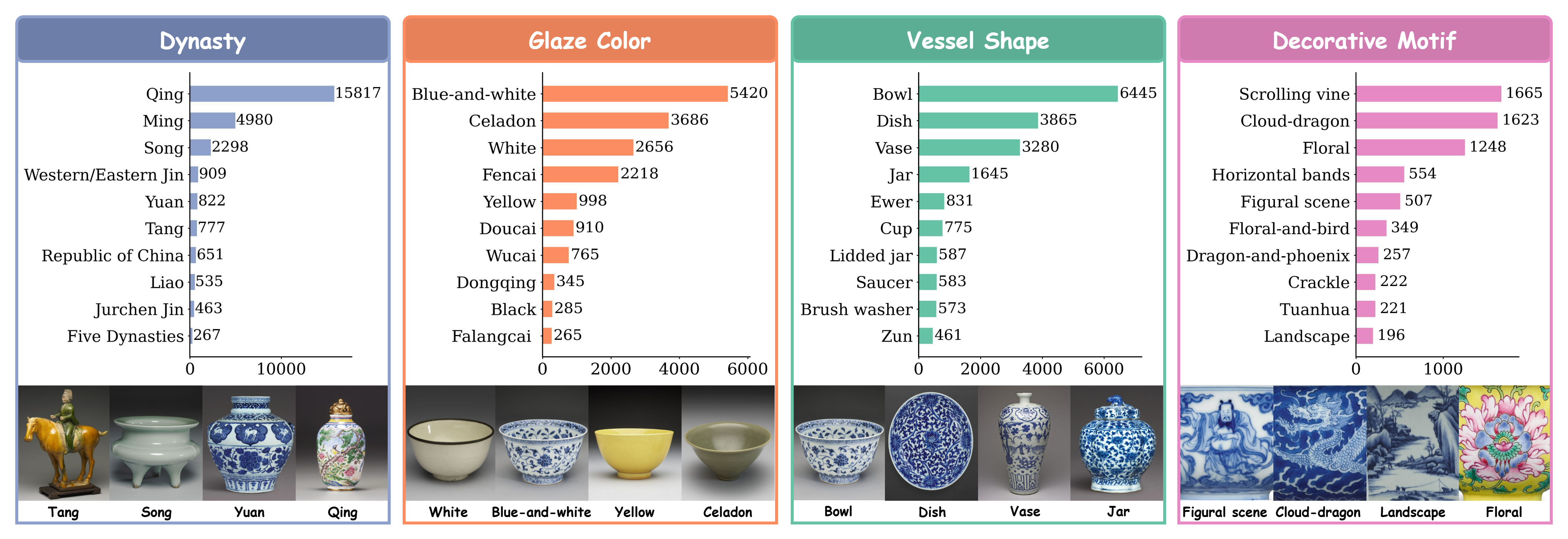}
   \caption{
        \textbf{Visualization of the four key attributes in porcelain connoisseurship.}
        Shown are the distributions of dynasty, glaze color, vessel shape, and decorative motif in the raw porcelain dataset. 
        For each attribute, the top 10 most frequent categories are presented.
        }
   \label{fig:dataset_distribution}
\end{figure*}

\subsection{CiQi-VQA dataset} \label{subsec:vqa_ds}

\medskip
\noindent\textbf{Raw Data Collection.}
We first collect raw antique Chinese porcelains from multiple publicly accessible sources, including web-based searches, open-access digital museum collections, and digitized scholarly books. From these sources, we curate a dataset of $29{,}596$ unique specimens spanning $38$ dynasties, $42$ reign periods, $246$ glaze colors, $248$ decorative motif categories, and $158$ vessel shapes. To the best of our knowledge, this is the most comprehensive dataset for porcelain appreciation currently available.\footnote{A comparison with existing porcelain-related datasets is provided in the supplementary materials.}
Each porcelain specimen is associated with at least one high-quality image and a standardized name that explicitly encodes \textbf{four key attributes}: \textit{dynasty}, \textit{glaze color}, \textit{vessel shape}, and \textit{decorative motif}. In addition, a portion of the specimens further specifies \textbf{two additional attributes}—\textit{reign period} and \textit{kiln origin}. A complete example with all six attributes would be: “Qing Dynasty, Kangxi period (1662–1722 CE), Jingdezhen kiln, Blue-and-white, Cloud-and-dragon motif, Bowl (\zhtext{清康熙\ 景德镇青花云龙纹碗}).” Figure~\ref{fig:dataset_distribution} visualizes the distribution across the four key attribute dimensions, showing the top-$10$ most frequent categories in each. 

From this collection, we later select $775$ pieces to construct CiQi-Bench for evaluation, while the remaining $28{,}821$ pieces are used to build the CiQi-VQA training set.

\medskip
\noindent\textbf{Metadata Enrichment.}

% We further extract and clean the descriptive texts related to connoisseurship details from the raw porcelain data sources and aligned them with their corresponding raw porcelain specimens. For specimens that lacked such detailed descriptions and only had standardized names, we invite human experts to compose informative narrative descriptions. Then, we feed the raw images, standardized porcelain names, and detailed descriptive texts (from raw data and human experts) into MLLMs to produce a polished, model-aligned connoisseurship description for each specimen, specifically structured around the six connoisseurship attributes.
We further extract and clean descriptive texts related to porcelain connoisseurship from the raw sources and align them with each specimen. However, a substantial portion of the collected porcelains only provides standardized names without detailed narrative descriptions. To address this issue, we invite human experts to participate in the metadata enrichment process. Specifically, for 61.18\% of the specimens that lack detailed descriptions, the expert team composes complementary connoisseurship descriptions based on the specimen images and the available source metadata. In addition, to improve the reliability of the dataset, the experts review and correct the naming accuracy of the standardized names for all collected specimens, ensuring that the encoded attribute information is consistent and properly formatted. The expert team is led by a senior researcher with more than 20 years of experience in porcelain identification and connoisseurship research, and includes four graduate students from related disciplines who contribute to description completion and naming verification under the leader's supervision. Finally, we feed the standardized names, enriched descriptive texts, and raw images into MLLMs to produce a polished connoisseurship description for each specimen, which is structured in six paragraphs corresponding to six connoisseurship attributes.

\begin{figure}[t]
  \centering
   \includegraphics[width=1\linewidth]{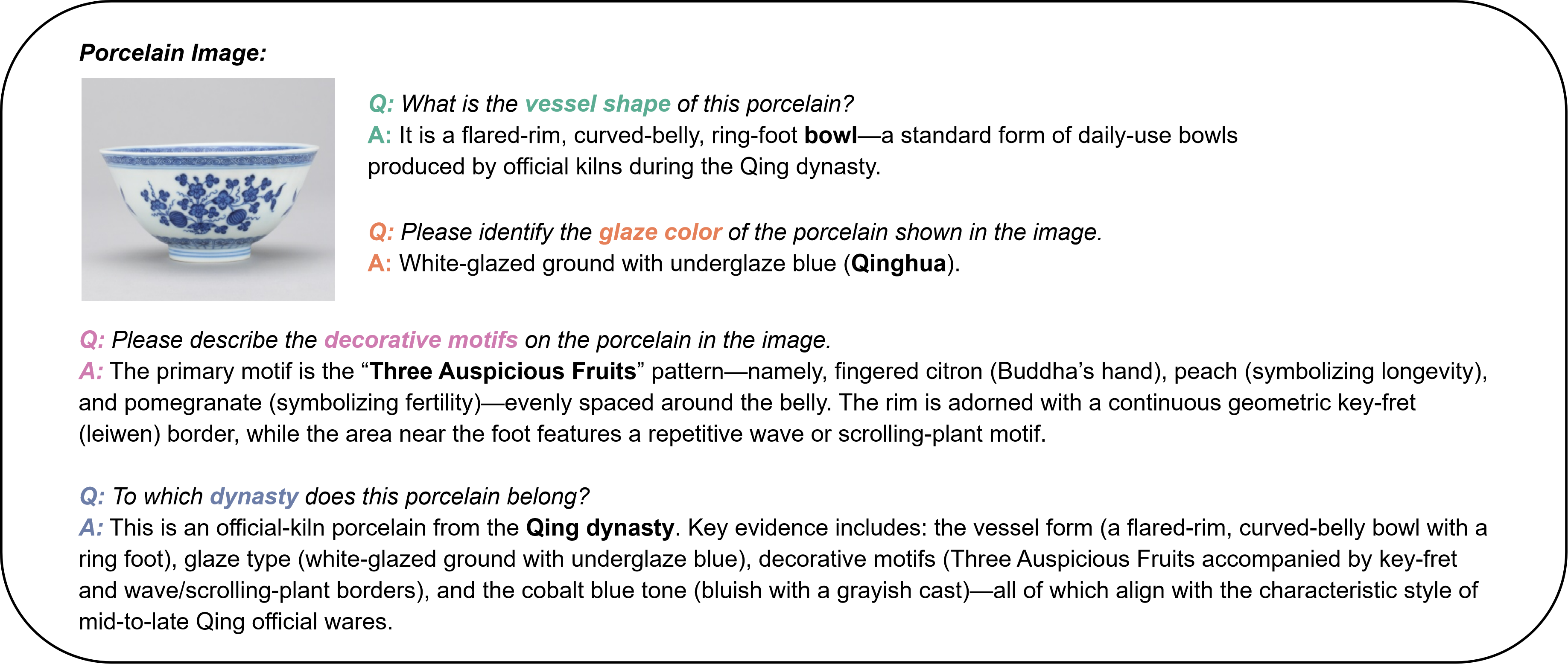}
   \caption{
    \textbf{Examples from CiQi-VQA dataset.}
    The figure illustrates annotations across four key attributes: dynasty, glaze color, decorative motif and vessel shape.
    }
   \label{fig:vqa_example}
\end{figure}

\medskip
\noindent\textbf{VQA Data Generation.}

We leverage MLLMs to generate specialized VQA pairs targeting the four key attributes, \textit{dynasty}, \textit{glaze color}, \textit{decorative motif}, and \textit{vessel shape}, as illustrated in Figure~\ref{fig:vqa_example}. We focus on these four rather than the full six benchmark attributes because they are most central to connoisseurship; the remaining two are treated as more advanced refinements and are emphasized at the RL stage.
For each porcelain specimen, the holistic description is also converted into a VQA format, giving five VQA training samples per item.

To further diversify linguistic expression, we adopt a lightweight augmentation strategy instead of multi-epoch training on identical samples. For each of the five VQA samples, we prompt the LLM to generate four additional variants that preserve the semantics but differ in phrasing and style. The final CiQi\mbox{-}VQA training set thus contains $20$ stylistically diverse yet semantically consistent VQA samples per specimen\footnote{For a small subset of specimens (e.g., monochrome wares), only $15$ questions are available because these objects inherently lack decorative motifs.}, and we perform SFT for a single epoch on this augmented dataset.

\begin{table}[t]
\centering
\caption{Overall statistics of CiQi-VQA dataset and CiQi-Bench.}
\label{tab:dataset_stat}
\small
\begingroup
\setlength{\aboverulesep}{1pt}
\setlength{\belowrulesep}{1pt}
\setlength{\cmidrulesep}{1pt}
\resizebox{\textwidth}{!}{%
\begin{tabular}{@{}lccccc@{}}
\toprule
 & \multirow{2}{*}{Porcelains} & \multirow{2}{*}{Images} &
\multicolumn{2}{c}{Questions} & \multirow{2}{*}{Attributes} \\[-0.3ex]
\cmidrule(lr){4-5}
 & & & VQA & Multiple-choice & \\
\midrule
SFT          & 28,821   & 50,675 & 557,165 & ---     & \\
RL$^*$       & 10,275   & 10,275 & 10,275  & ---     & dynasty, reign, kiln, color, motif, shape  \\
Evaluation   & 775      & 878    & 775     & 5,425   & \\
\midrule
Total        & 29,596   & 51,553 & 557,940 & 5,425   & \makecell[c]{dynasty, reign, kiln, color, motif, shape} \\
\bottomrule
\end{tabular}
}
\endgroup
\begin{minipage}{\textwidth}
\small
\centering
% \raggedright
$^*$ The raw porcelain data used for RL is a subset of that used for SFT.
\end{minipage}
\end{table}

\subsection{CiQi-Bench}

% 1.选择题，如何生成的，格式是怎样。2.LLM去评分，prompt（放在附录）。
% benchmark部分，我们使用了705件瓷器样本，并且设计了两种评价方式。第一种是使用生成选择题的方法，针对这705件瓷器样本的4个基本维度（朝代/器型/釉色/纹饰内容）和2个额外维度（皇帝年号/窑口）以及1个整体命名，对每件瓷器生成7道选择题，使用模型的回答正确率作为指标。第二种方式是直接让模型输出对于瓷器样本的整体描述，使用LLM对于模型回复的整体描述和瓷器整体描述的ground truth的相似性进行打分，同样输出对4个基本维度和2个额外维度的分别打分。

% \begin{figure*}[t]
%   \centering
%    \includegraphics[width=0.95\linewidth]{./sec/figures/agent_framework.png}

%    \caption{
%     \textbf{Framework of the CiQi-Agent.}
%     The agent integrates visual zoom-in, image/text retrieval tools within a two-phase training pipeline.
%     Supervised fine-tuning establishes tool-calling skills and porcelain connoisseurship knowledge, while reinforcement learning with an LLM-as-a-Judge refines accuracy and strategic tool-calling. 
%     }
%    \label{fig:agent_framework}
% \end{figure*}

For the benchmark, we curated a set of $775$ porcelain specimens and designed two evaluation protocols.

\medskip
\noindent\textbf{Multiple-Choice Questions.} The first protocol adopts a \textit{multiple-choice} format. For each specimen, we construct seven questions covering the four key attributes, two additional attributes (reign period and kiln origin), and the full standardized name. We use GPT-5 to automatically generate the multiple-choice questions: given the image and the ground-truth annotation as input, the model is instructed to produce plausible yet challenging distractor options. Model performance is then quantified by the answer accuracy over all questions. 

\medskip
\noindent\textbf{Free-Form Questions.} The second protocol focuses on \textit{free-form generation}. In this setting, the model is prompted to produce a holistic textual description of each porcelain specimen. An LLM-based evaluator is subsequently employed to compare the generated description with the ground-truth text and assign six separate similarity scores---one for each of the four key attributes and the two additional attributes.\footnote{The specific MLLMs and prompt templates used during the construction of our dataset and benchmark are provided in the supplementary materials.
}

Table~\ref{tab:dataset_stat} summarizes the overall statistics of the porcelain dataset and benchmark. The raw porcelain data used for RL is a subset of that used for SFT.

\section{Training Framework of CiQi-Agent}\label{sec:agent}

\begin{figure*}[t]
  \centering
   \includegraphics[width=1\linewidth]{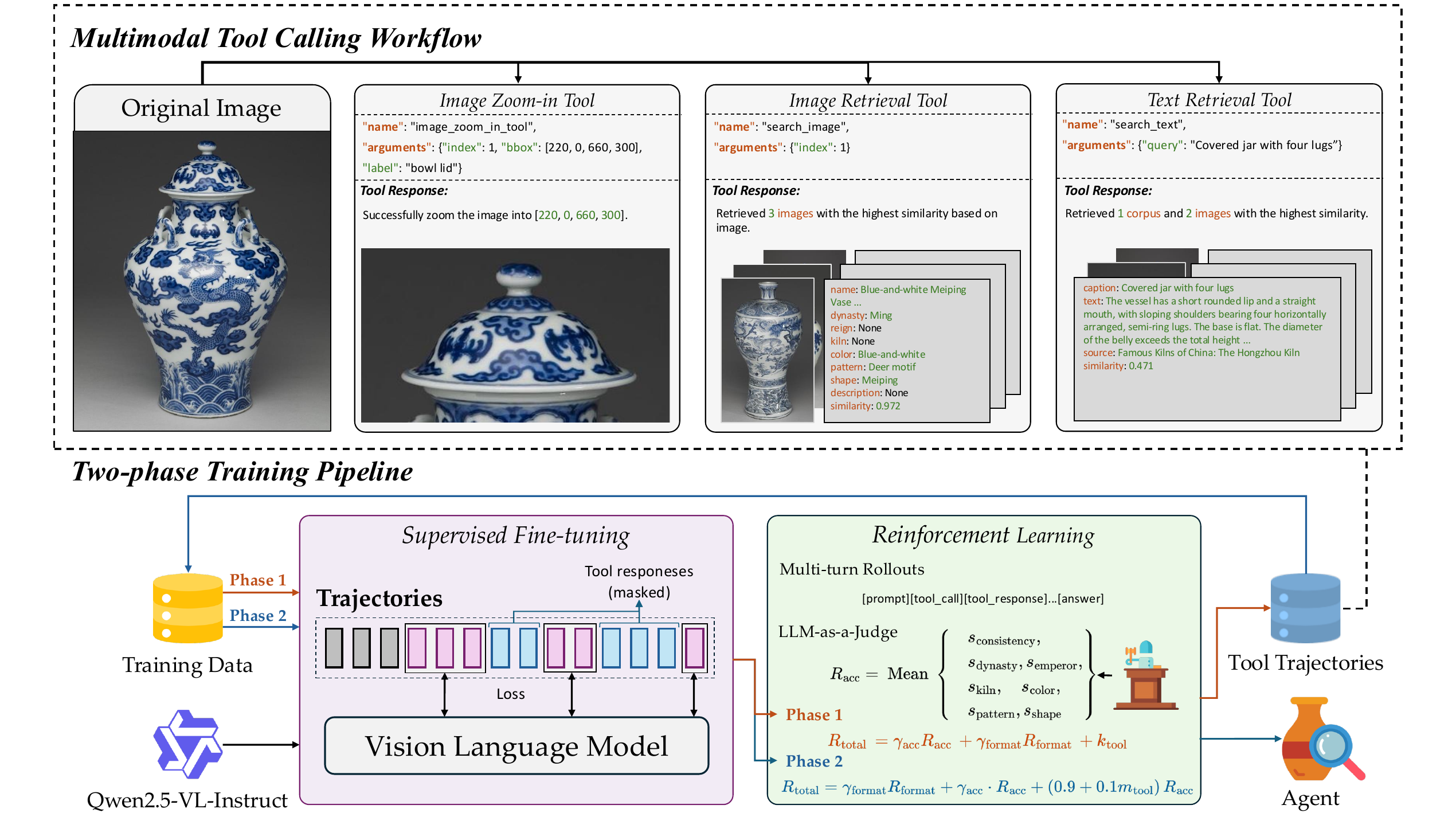}

   \caption{
    \textbf{Framework of the CiQi-Agent.}
    The agent integrates visual zoom-in, image/text retrieval tools within a two-phase training pipeline.
    Supervised fine-tuning establishes tool-calling skills and porcelain connoisseurship knowledge, while reinforcement learning with an LLM-as-a-Judge refines accuracy and strategic tool-calling. 
    }
   \label{fig:agent_framework}
\end{figure*}

In this section, we present the overall architecture of our proposed CiQi-Agent, which emulates the reasoning process of human experts by combining visual perception, retrieval-augmented knowledge access with reinforcement-driven tool-calling. As illustrated in Figure~\ref{fig:agent_framework}, the agent autonomously analyzes visual details, retrieves multimodal evidence, and generates context-aware judgments through the coordinated use of specialized tools.

% \begin{figure*}[htbp]
%   \centering
%    \includegraphics[width=0.95\linewidth]{./sec/figures/agent_framework.png}

%    \caption{
%     \textbf{Framework of the CiQi-Agent.}
%     The agent integrates visual zoom-in, image/text retrieval tools within a two-phase training pipeline.
%     Supervised fine-tuning establishes tool-calling skills and porcelain connoisseurship knowledge, while reinforcement learning with an LLM-as-a-Judge refines accuracy and strategic tool-calling. 
%     }
%    \label{fig:agent_framework}
% \end{figure*}

\subsection{Tool Design}\label{subsec:tool_design}
% 本章主要介绍我们为agent设计的可以调用的工具，来辅助瓷器鉴定agent完成任务。工具分为两类，一类是视觉工具，如图像放大等；一类是检索工具，如图像检索，文本检索等。
To enable more flexible and interpretable reasoning, a suite of external tools is provided, which can be autonomously selected and executed by CiQi-Agent during inference.
Each tool invocation is encapsulated within a standardized \texttt{<tool\_call></tool\_call>} tag pair, allowing the agent to issue structured commands and integrate tool outputs into its reasoning context in a consistent format.
These tools are organized into two categories, \textit{vision tool} and \textit{retrieval tool}, which respectively serve the purposes of perceptual enhancement and knowledge acquisition, as illustrated in Figure~\ref{fig:agent_framework}.

% 视觉工具方面，我们设计了一个图像放大工具，来帮助模型获得想关注的图像区域的细节。通过传入要检索的图像在上下文中的序号与想要截取的图像区域的左上角与右下角的顶点坐标，模型可以获得该图像区域的细节图像。这个细节图像会作为工具调用的结果传回模型上下文中来进行推理。
% 由于一张图像的有效特征在上下文中的占比并不算大，我们采用图像放大工具的好处在于压缩了高清图像在模型推理上下文的长度，减少模型推理的显存消耗，同时使得模型的注意力可以更好地向细节区域集中。
\medskip
\noindent\textbf{Vision Tool.}
The \textit{image zoom-in tool} enables the agent to autonomously focus on visually salient regions that are potentially informative for porcelain connoisseurship.
Instead of relying on predefined or user-specified coordinates, the agent first analyzes the global image context to infer which local areas merit closer examination—such as decorative motifs, glaze textures, or inscription details.
It then dynamically predicts the corresponding bounding-box parameters and extracts high-resolution visual patches from those regions.
The resulting sub-images are subsequently reintegrated into the multimodal reasoning context, where they serve as fine-grained perceptual evidence for the ongoing analysis.

% 检索工具方面，我们分别设计了图像检索工具和文本检索工具，帮助模型从瓷器数据库中检索出与当前任务相关的图像和文本信息，来辅助其进行推理。
% 通过这些调用，模型可以建立基于证据的推理，类似于咨询数字档案馆或博物馆数据库。
% 对于图像检索工具，通过传入要检索的图像在上下文中的序号，模型可以获得与该图像相似度最高的数个图像及其元数据；对于文本检索工具，通过传入要检索的文本，模型可以获得与该文本相似度最高的数个文本及其元数据。
% 这两个工具所检索的数据库是一个融合了高清标准器件瓷器图像与瓷器鉴定相关文本的图文融合RAG数据库。为了对图像与文本进行有效的特征表征，我们分别采用了CLIP模型和文本embedding模型来分别对数据库内的图文信息进行编码。
% 在模型对图像进行检索时，会先通过CLIP模型对图像进行编码，然后通过余弦相似度计算与数据库内所有图像的相似度，并返回相似度最高的向量的元数据；
% 在模型对文字进行检索时，会将文字分别转换为clip模型和text embedding模型的特征向量，然后通过余弦相似度计算与数据库内所有文字的相似度，并返回相似度最高的向量的元数据。
\medskip
\noindent\textbf{Retrieval Tools.}
Two retrieval tools are provided: an \textit{image retrieval tool} and a \textit{text retrieval tool}, enabling the agent to access evidence from a multimodal porcelain database autonomously.
Both tools operate on a unified RAG framework that integrates high-resolution images and textual knowledge related to porcelain connoisseurship.

For \textit{image retrieval}, the query image is encoded by the CLIP encoder~\cite{yang2023chineseclipcontrastivevisionlanguage}, and cosine similarity is computed against all image embeddings in the database.
The system returns top-$k$ similar entries and their metadata, which are reinserted into the reasoning context as visual evidence.
For text retrieval, the query text is encoded by both a CLIP encoder and a text-embedding model~\cite{chen2024bgem3embeddingmultilingualmultifunctionality}, yielding two vectors in distinct semantic spaces.
Each is matched with its corresponding index via cosine similarity, and the results are fused to identify the most relevant records by a parameter $\alpha$, which controls the relative contribution of each model to the final retrieval set.
The retrieved metadata, such as descriptions, provenance, or linked images, are then supplied to the agent as external knowledge.
This dual-space retrieval mechanism grounds the model’s reasoning in complementary visual and textual evidence, enhancing factual reliability and interpretability.

\subsection{Two-phase Training Pipeline}\label{subsec:overview}
Given the base model's initial lack of tool-calling competence and the absence of expert-annotated procedural trajectories, the acquisition of tool-calling skills is relegated to unguided exploration in the RL phase. This deficiency leads to a sample-inefficient process that constrains the attainable performance. Consequently, we employ a \textit{two-phase} training pipeline, with each phase consisting of an SFT stage and a subsequent RL stage.

\medskip
\noindent\textbf{Phase I: Foundational Competence.}
The first phase aims to establish general connoisseurship knowledge and tool-calling ability. 
During the SFT, the model is trained on the CiQi-VQA dataset (Sec.~\ref{subsec:vqa_ds}), augmented with instruction-following data~\cite{taori2023alpaca} to retain general reasoning capability and with $10{,}575$ everyday porcelain samples to reduce overfitting.
The subsequent RL stage employs \textit{Group Relative Policy Optimization (GRPO)}~\cite{shao2024deepseekmath}, and a strong tool-calling reward encourages the model to quickly learn the mechanics of tool-calling, yielding a model that generates synthetic tool-calling trajectories for \textit{Phase II}.

\medskip
\noindent\textbf{Phase II: Strategic Competence.}
The second phase focuses on refining the agent’s ability to integrate tool-calling with higher-level reasoning. 
The synthetic trajectories from \textit{Phase~I} are merged back into the SFT corpus, allowing the model to internalize both reasoning and tool-calling patterns. 
The subsequent RL stage employs a redesigned reward that explicitly links tool-calling rewards to connoisseurship accuracy, encouraging the model to improve connoisseurship performance through more effective and purposeful tool-calling.
Through this two-phase curriculum, the agent progressively acquires domain-specific expertise required for porcelain connoisseurship while learning to call tools to enhance perception and reasoning.

\subsection{Reward Design}
\label{subsec:reward_design}
During RL, the overall reward function is a weighted average of \textit{format reward} $R_{\text{format}}$, \textit{accuracy reward} $R_{\text{acc}}$, and \textit{tool-calling reward} $R_{\text{tool}}$:
\begin{equation*}
    R = \gamma_\text{format} \cdot R_{\text{format}} + \gamma_{\text{acc}} \cdot R_{\text{acc}} + R_{\text{tool}},
\end{equation*}
where weight parameters $\gamma_\text{format}$ and $\gamma_{\text{acc}}$ control the relative importance of the format and accuracy rewards.

\noindent\textbf{Format Reward.}
The format reward $R_{\text{format}} \in \{0, -1\}$ enforces compliance with the prescribed output and tool-calling syntax. Specifically, the agent receives $R_{\text{format}} = 0$ if both the response and tool calls strictly follow the required format; otherwise, it incurs a penalty of $R_{\text{format}} = -1$.

\medskip
\noindent\textbf{Accuracy Reward.} 
The accuracy reward $R_{\text{acc}}$ measures how accurately the model names the porcelain. It is composed of six attribute scores: \textit{dynasty}, \textit{reign period}, \textit{kiln origin}, \textit{glaze color}, \textit{decorative motif}, and \textit{vessel shape}, plus a \textit{consistency} score that checks whether the output conforms to the required format. Each score $s_i \in [0, 1]$. If the ground-truth does not contain a certain attribute, that attribute is excluded from the average. Formally, let $\mathcal{M}$ denote the set of attribute indices present in the ground-truth for a given sample; then $R_{\text{acc}} = |\mathcal{M}|^{-1} \sum_{i \in \mathcal{M}} s_i$.
The evaluation is performed by the LLM-as-a-Judge described in Sec.~\ref{subsec:llm-as-a-judge}.

\medskip
\noindent\textbf{Tool-calling Reward.}
The tool-calling reward $R_{\text{tool}}$ is designed to guide the agent’s learning of tool-calling behavior and its effective use in improving connoisseurship performance.
\begin{itemize}
    \item During the \textit{first-phase} RL training, the primary objective is to help the model quickly acquire the ability to call tools appropriately.
For each rollout, if the model invokes $k_{\text{tool}}$ tools, it receives a corresponding reward $R_\text{tool}=k_{\text{tool}}$, proportional to the number of successful tool-calls, encouraging early mastery of tool-calling behavior.
    \item During the \textit{second-phase} RL training, the reward scheme shifts from quantity to quality.
To promote strategic and meaningful tool-calling, the agent is rewarded only when the invocation of tools contributes to higher connoisseurship accuracy.
Specifically, if the model invokes $m_{\text{tool}}$ \textit{distinct} tools during a rollout, it receives a scaled tool-calling reward
$R_{\text{tool}} = (0.9 + 0.1 m_{\text{tool}})\, R_{\text{acc}}$ 
while rollouts without any tool invocation receive no tool-calling reward.
\end{itemize}
This design incentivizes the model to explore tool-calling strategies that genuinely enhance identification accuracy rather than indiscriminate calling of tools.

\subsection{LLM-as-a-Judge}
\label{subsec:llm-as-a-judge}
Although porcelain naming follows a standardized attribute structure, linguistic realizations of each attribute exhibit substantial lexical variation while preserving semantic equivalence. As a result, exact string matching or embedding-based similarity metrics fail to reliably assess correctness, particularly when fine-grained, domain-specific terminology is involved. 
To address this challenge, we instead adopt an \emph{LLM-as-a-Judge} approach: during RL training and evaluation, the agent is required to output a standardized naming enclosed within a predefined \texttt{<answer></answer>} tag pair. The content inside these tags, along with the corresponding ground-truth porcelain name, is submitted to an LLM-based evaluator. The evaluator is instructed to rate each attribute according to a structured scoring rubric, yielding individual scores $s_i \in [0, 1]$, where higher values indicate greater accuracy or stylistic consistency. The detailed prompt is provided in~\ref{supp_subsec:prompt_llm-as-a-judge_eval}--\ref{supp_subsec:prompt_llm-as-a-judge_train} of the Supplementary Material.

To validate the reliability of LLM-as-a-Judge, we run the SFT model after first-phase training on CiQi-Bench to generate predictions and collect independent evaluations from domain experts using the same scoring rubric. Table~\ref{tab:llm_judge_correlation} reports the Pearson correlation coefficients and mean absolute errors (MAE) between expert scores and LLM-based scores across all six attributes. The consistently high correlations, together with the small MAE values, indicate strong alignment between the LLM-based evaluator and human expert judgment.

\begin{table}[t]
    \centering
    \footnotesize
    \caption{Pearson Correlation and MAE between human expert scores and LLM-as-a-Judge scores on CiQi-Bench. All values are reported to three decimal places.}
    \label{tab:llm_judge_correlation}
    {
    \begin{tabular*}{\textwidth}{@{\extracolsep{\fill}}lrrrrrr}
        \toprule
        & Dynasty & Reign period & Kiln & Glaze color & Motif & Shape \\
        \midrule
        Pearson $r$ & 0.995 & 1.000 & 0.979 & 0.958 & 0.938 & 0.859 \\
        MAE & 0.013 & 0.000 & 0.036 & 0.028 & 0.065 & 0.077 \\
        \bottomrule
    \end{tabular*}
    }
\end{table}

\section{Experiments}
% 在这一章节，我们设置了一系列的实验来评估我们提出的CiQi-Agent的性能。我们首先在CiQi-Bench上进行了一系列的实验，然后进行了一些ablation study来分析我们模型的各个部分对于性能的影响。

In this section, we conduct extensive experiments to evaluate the performance of our proposed CiQi-Agent. We first conduct experiments on the CiQi-Bench,  and then perform ablation studies to analyze the impact of each component on our model's performance.
\subsection{Experiment Setup}
\noindent\textbf{Baseline Configuration.}
We compare our method against a comprehensive set of state-of-the-art multimodal models, including both closed-source and open-source variants.
For closed-source models (GPT-5~\cite{openai2025gpt5}, GPT-4.1~\cite{openai2025gpt41}, GPT-4o~\cite{openai2024gpt4o}, OpenAI o3~\cite{openai2025introducingo3}, Gemini 2.5 Pro~\cite{gemini25pro}, and Claude Opus 4~\cite{anthropic_claude4_systemcard_2025}), we use their official APIs with the default inference settings, with the temperature set to $0.0$ to ensure deterministic outputs.
For open-source models, we select the largest released models for each model family (Qwen2.5-VL-72B-Instruct~\cite{qwen25vl}, GLM-4.5V~\cite{glm45v_2025}, InternVL3.5-241B-A28B-Flash~\cite{internvl35_2025}, Kimi-VL-A3B-Instruct~\cite{kimivl_2025}). Models are loaded using their official implementations and evaluated with temperature=$0.0$ to ensure reproducibility.

\medskip
\noindent\textbf{Our Method Configuration.}
Our agent is built upon Qwen2.5-VL-7B-Instruct, trained using the two-phase pipeline described in Sec.~\ref{sec:agent}, with Qwen2.5-72B-Instruct~\cite{qwen25} served as the LLM-as-a-Judge evaluator. The weight parameters are set as $\gamma_{\text{format}}=0.2$ and $\gamma_{\text{acc}}=1.0$, respectively. For retrieval, we construct a database consisting of $16{,}830$ images and $49{,}606$ texts from our CiQi-VQA dataset, which is strictly non-overlapping with both the RL training set and the CiQi-Bench to avoid information leakage and ensure fair assessment. See~\ref{supp_sec:detailed_config} in the Supplementary Material for the detailed configurations.

\medskip
\noindent\textbf{Evaluation Metrics.}
We report accuracy as the primary metric for multiple-choice questions, computed as the percentage of correctly answered questions across all seven dimensions: overall naming, dynasty, reign period, kiln site, glaze color, decorative motif, and vessel shape.
For free-form generation tasks, we employ Qwen2.5-72B-Instruct~\cite{qwen25} as the LLM-as-a-Judge evaluator.

\subsection{Main Results on CiQi-Bench}
Table~\ref{tab:performance-combined} summarizes the results on CiQi-Bench.
CiQi-Agent achieves \textbf{state-of-the-art performance} across all dimensions, validating the effectiveness of our two-phase training pipeline and tool-augmented reasoning framework.

On multiple-choice tasks, CiQi-Agent attains $85.2\%$ overall naming accuracy and an average of $81.5\%$ across all seven attributes, surpassing GPT-5 (average $75.8\%$) and Qwen2.5-VL-72B-Instruct (average $69.2\%$).
Notably, it achieves $77.6\%$ on dynasty, $70.3\%$ on reign period, and $81.8\%$ on kiln site---outperforming all baselines by substantial margins.
Performance gains are especially pronounced in visually grounded attributes, reaching $91.4\%$ on glaze color, $75.7\%$ on decorative motif, and $88.1\%$ on vessel shape.

For free-form generation, CiQi-Agent further demonstrates strong generalization, achieving $71.3\%$ on dynasty (vs.~$42.7\%$ for o3), $69.8\%$ on kiln site (vs.~$44.4\%$ for o3), and $85.4\%$ on glaze color (vs.~$75.8\%$ for Qwen2.5-VL-72B), with an average of $66.7\%$ across all six attributes (vs.~$43.0\%$ for Qwen2.5-VL-72B and $48.0\%$ for GPT-5).
Remarkably, despite having only 7B parameters, CiQi-Agent outperforms GPT-5 across all attributes and in both multiple-choice and free-form averages, demonstrating that tool-calling and domain-aligned training can effectively compensate for model scale.
% 值得指出的是，我们的模型在free-form上的优势远大于multiple-choice，这表明我们提出的CiQi-Agent在对瓷器的整体感知上具有非常强大的能力。特别是在Dynasty和Reign方面，我们的模型的表现远远高于其他模型，对于这两个属性的识别，依赖于对瓷器更为精准的细节观察，以及对不同朝代的瓷器风格和特征的准确识别。baseline模型往往较为保守，不能给出明确的时代答案，但是得益于视觉工具的使用，与大量的高质量瓷器样本的训练，我们的模型能够针对不同的瓷器样本，主动选择是否给出更为细致的年代判断，并且具有很高的正确率。
% CiQi-Agent’s gains are substantially larger on free-form than on multiple-choice tasks, underscoring strong holistic, evidence-grounded ceramic perception. The advantage is most pronounced for Dynasty and Reign, which hinge on fine-grained morphological cues and dynasty-specific stylistic signatures. Whereas baselines often hedge and avoid firm temporal attributions, our tool-augmented agent (zoom-in + retrieval), trained on a large curated ceramic corpus, learns when to commit to granular chronological judgments---and achieves markedly higher accuracy when doing so.
\begin{table*}[t]
    \centering
    \small
    \caption{Performance on CiQi-Bench (Multiple-choice \& Free-form). Bold indicates the best performance, and underline indicates the second-best performance.}
    \resizebox{\textwidth}{!}{%
    \begin{tabular}{lrrrrrrrr|rrrrrrr}
    \toprule
    \multicolumn{1}{c}{\multirow{2}[4]{*}{Model}} 
    & \multicolumn{8}{c}{Multiple-choice Accuracy (\%)} 
    & \multicolumn{7}{c}{Free-form Accuracy (\%)} \\
    \cmidrule(lr){2-9} \cmidrule(lr){10-16}
    & Dynasty & Reign & Kiln & Color & Motif & Shape & Naming & Average
    & Dynasty & Reign & Kiln & Color & Motif & Shape & Average \\
    \midrule
    % ---- Closed-source ----
    GPT-5~\cite{openai2025gpt5} & \underline{65.7} & 61.4 & \underline{79.6} & 86.5 & 69.3 & 83.8 & \underline{84.3} & \underline{75.8} 
          & 39.4 & 32.8 & 42.6 & 74.4 & \underline{35.3} & 63.9 & 48.0 \\
    GPT-4.1~\cite{openai2025gpt41} & 59.3 & \underline{68.3} & 71.1 & 85.0 & 62.2 & 81.8 & 77.9 & 72.2 
            & 36.7 & 27.2 & 29.0 & 67.5 & 27.6 & 60.1 & 41.3 \\
    GPT-4o~\cite{openai2024gpt4o} & 59.1 & 60.4 & 68.6 & \underline{89.2} & 70.1 & \underline{84.2} & 82.1 & 73.4 
           & 26.9 & 13.4 & 15.1 & 53.9 & 21.1 & 47.6 & 29.7 \\
    o3~\cite{openai2025introducingo3} & 57.6 & 57.4 & 72.2 & 82.6 & 62.4 & 76.8 & 76.6 & 69.4 
       & \underline{42.7} & \underline{36.6} & 44.4 & 74.2 & 33.1 & 62.1 & \underline{48.8} \\
    Gemini 2.5 Pro~\cite{gemini25pro} & 54.4 & 57.4 & 68.0 & 65.2 & 58.5 & 64.2 & 82.5 & 64.3 
                   & \underline{48.1} & 34.9 & \underline{48.4} & 50.2 & 16.1 & 39.0 & 39.5 \\
    Claude Opus 4~\cite{anthropic_claude4_systemcard_2025} & 54.8 & 40.6 & 65.3 & 74.9 & 59.0 & 75.2 & 69.3 & 62.7 
                  & 36.8 & 10.3 & 22.0 & 64.1 & 25.1 & 59.1 & 36.2 \\
    \midrule
    % ---- Open-source ----
    % Qwen3-VL-235B-A22B-Instruct & \underline{69.1} & \underline{68.3} & {75.2} & \underline{90.5} & \underline{72.5} & \underline{85.3} & 83.9 & \underline{77.8} 
    %                             & \underline{48.9} & {35.9} & \underline{49.4} & \underline{77.2} & \underline{38.7} & \underline{69.2} & \underline{53.2} \\
    Qwen2.5-VL-72B-Instruct~\cite{qwen25vl} & 57.6 & 34.7 & 69.2 & 86.7 & \underline{71.7} & 84.1 & 80.3 & 69.2 
                            & 29.5 & 31.2 & 27.7 & \underline{75.8} & 31.0 & 62.6 & 43.0 \\
    GLM-4.5V (106B)~\cite{glm45v_2025} & 58.3 & 59.4 & 75.8 & 82.3 & 70.4 & 81.8 & 80.6 & 72.6 
             & 31.0 & 14.3 & 32.8 & 65.4 & 31.1 & \underline{65.2} & 39.9 \\
    InternVL3.5-241B-A28B-Flash~\cite{internvl35_2025} & 57.1 & 38.6 & 59.5 & 82.1 & 64.8 & 73.9 & 68.5 & 63.5 
                                & 42.4 & 31.6 & 36.9 & 52.6 & 19.6 & 41.5 & 37.4 \\
    Kimi-VL-A3B-Instruct (16B)~\cite{kimivl_2025} & 59.3 & 22.8 & 48.8 & 84.8 & 59.8 & 77.9 & 70.3 & 60.5 
                         & 17.3 & 23.7 & 16.2 & 69.5 & 26.5 & 61.3 & 35.7 \\
    \midrule
    % ---- Ours ----
    CiQi-Agent (Ours, 7B) & \textbf{77.6} & \textbf{70.3} & \textbf{81.8} & \textbf{91.4} & \textbf{75.7} & \textbf{88.1} & \textbf{85.2} & \textbf{81.5} 
                      & \textbf{71.3} & \textbf{49.1} & \textbf{69.8} & \textbf{85.4} & \textbf{49.7} & \textbf{75.0} & \textbf{66.7} \\
    \bottomrule
    \end{tabular}
    }
    \label{tab:performance-combined}
\end{table*}

% \begin{table*}[htbp]
%     \centering
%     \small
%     \caption{Ablation study on Qwen2.5-VL-7B variants. The table shows multiple-choice and free-form performance.}
%     \resizebox{\textwidth}{!}{%
%     \begin{tabular}{lrrrrrrr|rrrrrr}
%     \toprule
%     \multicolumn{1}{c}{\multirow{2}[4]{*}{Model}} & \multicolumn{7}{c}{Multiple-choice Accuracy (\%)} & \multicolumn{6}{c}{Free-form Accuracy (\%)} \\
%     \cmidrule{2-8} \cmidrule{9-14}
%     & Overall Naming & Dynasty & Reign & Kiln & Color & Motif & Shape & Dynasty & Reign & Kiln & Color & Motif & Shape \\
%     \midrule
%     Qwen2.5-VL-7B (Base) & 0.6904 & 0.5400 & 0.6040 & 0.5510 & 0.8364 & 0.6984 & 0.7956 & 0.2034 & 0.2218 & 0.1549 & 0.7059 & 0.2844 & 0.6136 \\
%     Base + SFT & 0.8187 & 0.6547 & 0.5347 & 0.7713 & 0.9261 & 0.7275 & 0.8823 & 0.6460 & 0.4515 & 0.5655 & 0.8163 & 0.3584 & 0.7088 \\
%     RL (w/o tools) &  &  &  &  &  &  &  &  &  &  &  &  &  \\
%     RL from V54 step 784 & 0.8420 & 0.6920 & 0.6634 & 0.5978 & 0.9103 & 0.7857 & 0.7671 & 0.6989 & 0.4248 & 0.5108 & 0.8383 & 0.4179 & 0.7372 \\
%     Ours & 0.8536 & 0.8027 & 0.7030 & 0.8402 & 0.9142 & 0.7910 & 0.8978 & 0.7199 & 0.5252 & 0.6803 & 0.8450 & 0.5072 & 0.7619 \\
%     \bottomrule
%     \end{tabular}%
%     }
%     \label{tab:ablation}
% \end{table*}

\subsection{Ablation Study}

\medskip
\noindent\textbf{Training stages.}
To understand how each training stage contributes to the final performance, we perform an ablation study based on Qwen2.5-VL-7B-Instruct, as shown in Table~\ref{tab:qwen-ablation-arrows}.

% The models employed in our ablation study include: (1) the base model, (2) the base model with SFT, (3) the base model with SFT and RL without tool-calling, (4) the base model with SFT and RL with the vision tool enabled, and (5) the base model with SFT and RL equipped with all available tools.
% Since the tool-use trajectories are generated using the \textit{Phase I} model, Model~(2) in the ablation corresponds to the \textit{Phase II} model obtained via SFT. Model~(3) to~(5) are built on the \textit{Phase II} model through RL, where model (5) represents our CiQi-Agent.
We conduct an ablation study with the following three model variants:
(1) the base model;
(2) the SFT model fine-tuned on the base model, corresponding to the \textit{Phase II} model;
(3) the SFT+RL model, i.e., our full \textbf{CiQi-Agent}.
Since the tool-calling trajectories used for RL training are generated by the \textit{Phase I} model, Model (3) is built on Model (2).
% Model (2) serves as the initialization point for Models (3)–(5). 

\begin{table*}[ht]
    \centering
    \small
    \caption{Ablation study on Qwen2.5-VL-7B-Instruct variants. Arrows indicate improvement (↑) or decline (↓) relative to the previous variant, and bold indicates the best performance.}
    \resizebox{\textwidth}{!}{%
        \begin{tabular}{lrrrrrrrr|rrrrrrr}
        \toprule
        \multicolumn{1}{c}{\multirow{2}[4]{*}{Model}} 
        & \multicolumn{8}{c}{Multiple-choice Accuracy (\%)} 
        & \multicolumn{7}{c}{Free-form Accuracy (\%)} \\
        \cmidrule(lr){2-9} \cmidrule(lr){10-16}
        & Dynasty & Reign & Kiln\phantom{↑} & Color & Motif & Shape & Naming & Average
        & Dynasty & Reign & Kiln\phantom{↑} & Color & Motif & Shape & Average \\
        \midrule
        Qwen2.5-VL-7B-Instruct~\cite{qwen25vl} 
        & 54.0\phantom{ ↑} & 60.4\phantom{ ↑} & 55.1\phantom{ ↑} & 83.6\phantom{ ↑} & 69.8\phantom{ ↑} & 79.6\phantom{ ↑} & 69.0\phantom{ ↑} & 67.4\phantom{ ↑} 
        & 20.3\phantom{ ↑} & 22.2\phantom{ ↑} & 15.5\phantom{ ↑} & 70.6\phantom{ ↑} & 28.4\phantom{ ↑} & 61.4\phantom{ ↑} & 36.4\phantom{ ↑} \\
    
        + SFT 
        & 65.5 ↑ & 53.5 ↓ & 77.1 ↑ & \textbf{92.6} ↑ & 72.8 ↑ & \textbf{88.2} ↑ & 81.9 ↑ & 75.9 ↑ 
        & 64.6 ↑ & 45.2 ↑ & 56.6 ↑ & 81.6 ↑ & 35.8 ↑ & 70.9 ↑ & 59.1 ↑ \\
    
        + SFT + RL
        & \textbf{77.6} ↑ & \textbf{70.3} ↑ & \textbf{81.8} ↑ & 91.4 ↓ & \textbf{75.7} ↑ & 88.1 ↓ & \textbf{85.2} ↑ & \textbf{81.5} ↑ 
        & \textbf{71.3} ↑ & \textbf{49.1} ↑ & \textbf{69.8} ↑ & \textbf{85.4} ↑ & \textbf{49.7} ↑ & \textbf{75.0} ↑ & \textbf{66.7} ↑ \\
        \bottomrule
        \end{tabular}%
    }
    \label{tab:qwen-ablation-arrows}
\end{table*}

    % \begin{tabular}{lrrrrrrr|rrrrrr}
    % \toprule
    % \multicolumn{1}{c}{\multirow{2}[4]{*}{Model}} & \multicolumn{7}{c}{Multiple-choice Accuracy (\%)} & \multicolumn{6}{c}{Free-form Accuracy (\%)} \\
    % \cmidrule{2-8} \cmidrule{9-14}
    % & Overall Naming & Dynasty & Reign & Kiln & Color & Motif & Shape & Dynasty & Reign & Kiln & Color & Motif & Shape \\
    % \midrule
    % Qwen2.5-VL-7B (Base) & 0.6904\phantom{ ↑} & 0.5400\phantom{ ↑} & 0.6040\phantom{ ↑} & 0.5510\phantom{ ↑} & 0.8364\phantom{ ↑} & 0.6984\phantom{ ↑} & 0.7956\phantom{ ↑} & 0.2034\phantom{ ↑} & 0.2218\phantom{ ↑} & 0.1549\phantom{ ↑} & 0.7059\phantom{ ↑} & 0.2844\phantom{ ↑} & 0.6136\phantom{ ↑} \\
    % Base + SFT & 0.8187 ↑ & 0.6547 ↑ & 0.5347 ↓ & 0.7713 ↑ & 0.9261 ↑ & 0.7275 ↑ & 0.8823 ↑ & 0.6460 ↑ & 0.4515 ↑ & 0.5655 ↑ & 0.8163 ↑ & 0.3584 ↑ & 0.7088 ↑ \\
    % RL (w/o tools) &  &  &  &  &  &  &  &  &  &  &  &  &  \\
    % RL from V54 step 784 & 0.8420 ↑ & 0.6920 ↑ & 0.6634 ↑ & 0.5978 ↓ & 0.9103 ↓ & 0.7857 ↑ & 0.7671 ↓ & 0.6989 ↑ & 0.4248 ↓ & 0.5108 ↓ & 0.8383 ↑ & 0.4179 ↑ & 0.7372 ↑ \\
    % Ours & \textbf{0.8536} ↑ & \textbf{0.8027} ↑ & \textbf{0.7030} ↑ & \textbf{0.8402} ↑ & 0.9142 ↑ & \textbf{0.7910} ↑ & \textbf{0.8978} ↑ & \textbf{0.7199} ↑ & \textbf{0.5252} ↑ & \textbf{0.6803} ↑ & \textbf{0.8450} ↑ & \textbf{0.5072} ↑ & \textbf{0.7619} ↑ \\
    % \bottomrule
    % \end{tabular}%% 

\noindent\textit{Effect of SFT.} SFT substantially boosts performance: overall naming improves from $69.0\%$ to $81.9\%$, and the multiple-choice average rises from $67.4\%$ to $75.9\%$. On free-form evaluation, dynasty accuracy jumps from $20.3\%$ to $64.6\%$, and the average increases from $36.4\%$ to $59.1\%$, establishing core connoisseurship knowledge.

\noindent\textit{Effect of RL.} Adding RL aligns tool-calling with knowledge-grounded reasoning: kiln accuracy jumps from $67.2\%$ to $81.8\%$ and dynasty from $51.6\%$ to $77.6\%$, and all free-form attributes improve (e.g., kiln: $63.6\% \rightarrow 69.8\%$, dynasty: $66.8\% \rightarrow 71.3\%$). Correspondingly, the multiple-choice average increases from $69.7\%$ to $81.5\%$, and the free-form average from $62.3\%$ to $66.7\%$, yielding our final CiQi-Agent.

% \noindent\textbf{Effect of SFT.}
% SFT substantially boosts performance: overall naming improves from $69.0\%$ to $81.9\%$, and the multiple-choice average rises from $67.4\%$ to $75.9\%$.
% On free-form evaluation, dynasty accuracy jumps from $20.3\%$ to $64.6\%$, and the average increases from $36.4\%$ to $59.1\%$, establishing core connoisseurship knowledge.

% \medskip
% \noindent\textbf{Effect of RL with vision tool.}
% Introducing the \textit{vision tool} enables finer local perception and achieves $84.2\%$ overall naming (multiple-choice average $75.1\%$), but induces regressions on kiln and color relative to the SFT-only variant ($77.1\% \rightarrow 59.8\%$ for kiln, $92.6\% \rightarrow 91.0\%$ for color), suggesting that perception-focused tool alone is insufficient to stabilize historically grounded attributions, even though the free-form average slightly increases from $59.1\%$ to $60.5\%$.

% \medskip
% \noindent\textbf{Effect of RL with all tools.}
% Adding \textit{retrieval} on top of zoom-in aligns tool use with knowledge-grounded reasoning: kiln accuracy jumps from $67.2\%$ to $81.8\%$ and dynasty from $51.6\%$ to $77.6\%$, and all free-form attributes improve (e.g., kiln: $63.6\% \rightarrow 69.8\%$, dynasty: $66.8\% \rightarrow 71.3\%$).
% Correspondingly, the multiple-choice average increases from $69.7\%$ to $81.5\%$, and the free-form average from $62.3\%$ to $66.7\%$, yielding our final CiQi-Agent.
\begin{table*}[h]
    \centering
    \small
    \caption{Ablation study on GPT-5, Qwen2.5-VL-7B-Instruct and CiQi-Agent with different tool configurations. Bold indicates the best performance.}
    \resizebox{\textwidth}{!}{%
        \begin{tabular}{lrrrrrrrr}
        \toprule
        \multicolumn{1}{c}{\multirow{2}[4]{*}{Model}} 
        & \multicolumn{8}{c}{Multiple-choice Accuracy (\%)} \\
        \cmidrule(lr){2-9}
        & Dynasty & Reign & Kiln & Color & Motif & Shape & Naming & Average \\
        \midrule
        GPT-5~\cite{openai2025gpt5}
        & 65.7 & 61.4 & 79.6 & 86.5 & 69.3 & 83.8 & 84.3 & 75.8 \\
    
        + vision tool 
        & 62.4 & 68.3 & 80.4 & 86.9 & 67.7 & 83.2 & 85.0 & 76.3 \\
    
        + retrieval tools 
        & 54.9 & 58.4 & 69.4 & 85.5 & 65.6 & 81.4 & 80.2 & 70.8 \\
    
        + all tools 
        & 55.7 & 58.4 & 68.3 & 85.8 & 66.7 & 82.3 & 78.6 & 70.8 \\
        \midrule
        Qwen2.5-VL-7B-Instruct~\cite{qwen25vl} 
        & 54.0 & 60.4 & 55.1 & 83.6 & 69.8 & 79.6 & 69.0 & 67.4 \\
    
        + vision tool 
        & 32.7 & 51.5 & 31.4 & 55.0 & 44.4 & 54.7 & 60.1 & 47.1 \\
    
        + retrieval tools 
        & 33.6 & 54.5 & 35.5 & 55.4 & 45.2 & 55.6 & 60.6 & 48.6 \\
    
        + all tools 
        & 32.3 & 53.5 & 36.9 & 54.1 & 45.2 & 55.4 & 59.3 & 48.1 \\
        \midrule
        CiQi-Agent (Ours, 7B) (w/o tools)
        & 51.6 & 32.7 & 67.2 & \textbf{92.7} & 73.3 & 86.2 & 84.3 & 69.7 \\
    
        + vision tool 
        & 69.2 & 66.3 & 59.8 & 91.0 & \textbf{78.6} & 76.7 & 84.2 & 75.1 \\

        + retrieval tools 
        & 68.7 & 59.4 & 81.0 & 88.9 & 75.4 & 83.3 & 84.3 & 77.3 \\
    
        + all tools
        & \textbf{77.6} & \textbf{70.3} & \textbf{81.8} & 91.4 & 75.7 & \textbf{88.1} & \textbf{85.2} & \textbf{81.5} \\    
        \bottomrule
        \end{tabular}%
    }
    \label{tab:gpt-qwen-tools-ablation}
\end{table*}

\medskip
\noindent\textbf{Tool Configurations.}
To understand how each tool contributes to the final performance, we perform an ablation study on GPT-5, Qwen2.5-VL-7B-Instruct, and our CiQi-Agent, as shown in Table~\ref{tab:gpt-qwen-tools-ablation}.

\noindent\textit{Tool on Base Models.} Tools do not consistently benefit general-purpose base models. (e.g., GPT-5’s average: $75.8\% \rightarrow 76.3\%/70.8\%/70.8\%$, Qwen2.5-VL-7B-Instruct's average: $67.4\% \rightarrow 47.1\%/48.6\%/48.1\%$). A plausible explanation is that the returned evidence (retrieved text or visual cues) is domain-specific and may require porcelain-appraisal knowledge to interpret and reconcile, which these base models do not reliably exhibit. In contrast, CiQi-Agent shows consistent gains (e.g., average: $69.7\%/75.1\%/77.3\% \rightarrow 81.5\%$), suggesting that models adapted with porcelain-domain expertise can more effectively interpret tool-provided evidence and translate it into improved performance.

\noindent\textit{Tool Combination on CiQi-Agent.} Combining tools is more effective than using a single tool for CiQi-Agent. The best average performance is achieved when vision and retrieval tools are enabled together (average: $81.5\%$), which supports the view that the two tools provide complementary information that is most useful when integrated.

Overall, the ablation results reveal a clear division of labor among training stages and tools.
SFT provides a comprehensive uplift of the model’s domain knowledge and yields consistent improvements across all evaluations.
Building on this foundation, RL with the vision tool strengthens the model’s ability to capture fine-grained visual cues, which is particularly beneficial for visually grounded attributes such as motif.
When multimodal retrieval tools are further incorporated, the agent gains the ability to compare input images against external porcelain exemplars, significantly improving history-based attributions, including dynasty, reign, and shape.
% These complementary contributions together lead to the final CiQi-Agent achieving the best overall performance on CiQi-Bench.

\section{Conclusion}
In this work, we present CiQi-Agent, a domain-specific multimodal agent for antique Chinese porcelain connoisseurship. We build CiQi-VQA, a large-scale dataset of expert-curated porcelain images and question–answer pairs, and CiQi-Bench, an expert-aligned benchmark that evaluates six connoisseurship attributes. For the training framework, we design a two-phase training pipeline that combines SFT, RL, and tool-augmented reasoning to align tool-calling with domain expertise. CiQi-Agent integrates visual zoom-in and image/text retrieval tools to perform fine-grained analysis with multimodal RAG. Extensive experiments show that CiQi-Agent significantly outperforms mainstream  MLLMs across all attributes, demonstrating the effectiveness of our dataset, benchmark, and training framework for cultural-heritage connoisseurship.

For future work, we plan to move beyond connoisseurship and tackle the more challenging task of authentication, i.e., distinguishing genuine antique porcelains from later imitations. In addition, CiQi-Agent represents a first step toward using MLLMs for cultural-heritage analysis; the same framework can be extended to build agents for other artifact types (e.g., ancient coins, calligraphy, paintings) or to develop a more general foundation model for cultural-heritage connoisseurship.
\section{Conclusion}
In this work, we present CiQi-Agent, a domain-specific multimodal agent for antique Chinese porcelain connoisseurship. We build CiQi-VQA, a large-scale dataset of expert-curated porcelain images and question–answer pairs, and CiQi-Bench, an expert-aligned benchmark that evaluates six connoisseurship attributes. For the training framework, we design a two-phase training pipeline that combines SFT, RL, and tool-augmented reasoning to align tool-calling with domain expertise. CiQi-Agent integrates visual zoom-in and image/text retrieval tools to perform fine-grained analysis with multimodal RAG. Extensive experiments show that CiQi-Agent significantly outperforms mainstream  MLLMs across all attributes, demonstrating the effectiveness of our dataset, benchmark, and training framework for cultural-heritage connoisseurship.

For future work, we plan to move beyond connoisseurship and tackle the more challenging task of authentication, i.e., distinguishing genuine antique porcelains from later imitations. In addition, CiQi-Agent represents a first step toward using MLLMs for cultural-heritage analysis; the same framework can be extended to build agents for other artifact types (e.g., ancient coins, calligraphy, paintings) or to develop a more general foundation model for cultural-heritage connoisseurship.

\bibliographystyle{plain}
\bibliography{main}

\clearpage
\onecolumn
\setcounter{page}{1}
% \maketitlesupplementary
\section*{Supplementary Material}
% \clearpage
% \setcounter{page}{1}
% \maketitle
% \section*{Supplementary Material}
% \def\confName{}
% \def\confYear{}

The supplementary material provides additional qualitative analyses, implementation details, and evaluation prompts for CiQi-Agent. It is organized as follows: Section~\ref{supp_sec:case_study} presents step-by-step case studies demonstrating how CiQi-Agent performs porcelain connoisseurship through multimodal reasoning and multi-stage tool invocation (image zoom-in, visual search, and textual retrieval); Section~\ref{supp_sec:prompt} documents all prompts used in our framework, including those for metadata enrichment, VQA data generation, multiple-choice option construction, and LLM-as-a-Judge in both training and evaluation; Section~\ref{supp_sec:comp_data}, compares the proposed CiQi-VQA dataset with existing porcelain-related datasets, highlighting its larger scale and finer-grained attribute coverage; finally, in Section~\ref{supp_sec:add_exp}, we conduct additional comparative experiments on the Hu et al.~\cite{hu2025integrating} dataset. These results further demonstrate the superior performance of our CiQi-Agent on datasets beyond CiQi-Bench.

\section{Case Study} \label{supp_sec:case_study}

\begin{tcolorbox}[title=Blue-and-White Flared-Rim Bowl with Figure Decoration{,} Kangxi Period{,} Qing Dynasty (\zhtext{清康熙{ }青花人物图撇口碗})]
\includegraphics[width=0.5\linewidth]{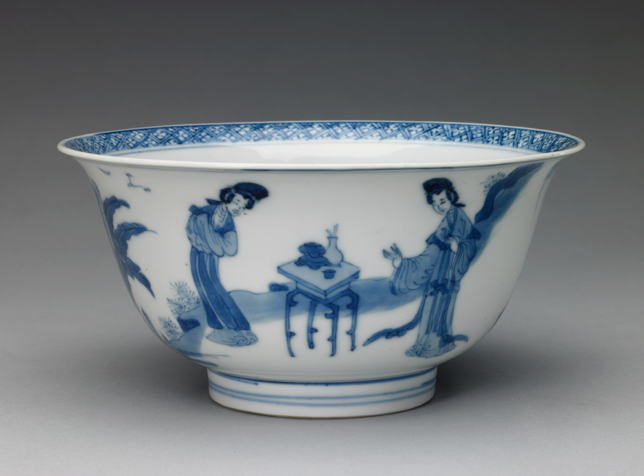}

\textbf{User:} Please provide a standardized name for this porcelain piece.

After your analysis, choose to either invoke a tool or respond directly, and place the final identification result within the \texttt{<answer></answer>} tags.

\textbf{Assistant:} This porcelain piece is a blue-and-white bowl with flared rim.
Its form features a flared mouth, rounded belly, and a ring foot, presenting an elegant and refined profile.
The exterior wall is decorated with blue-and-white figure scenes; the painting is meticulous, with vivid expressions, smooth garment lines, and background elements such as rocks and trees arranged with balanced density.
The cobalt blue appears pure and bright, displaying the typical “emerald-tinted blue” characteristic of Kangxi blue-and-white ware.
Along the inner rim is a band of blue-and-white diaper patterns, consistent with official kiln features of the Kangxi reign.

To further verify its dating and stylistic traits, I plan to invoke the search\_image tool to compare the bowl with stylistically similar pieces.
\begin{lstlisting}[style=casestyle]
<tool_call>
{"name": "search_image", "arguments": {"index": 1}}
</tool_call>
\end{lstlisting}

\textbf{User:} Successfully found the following content: 

\includegraphics[width=0.32\linewidth]{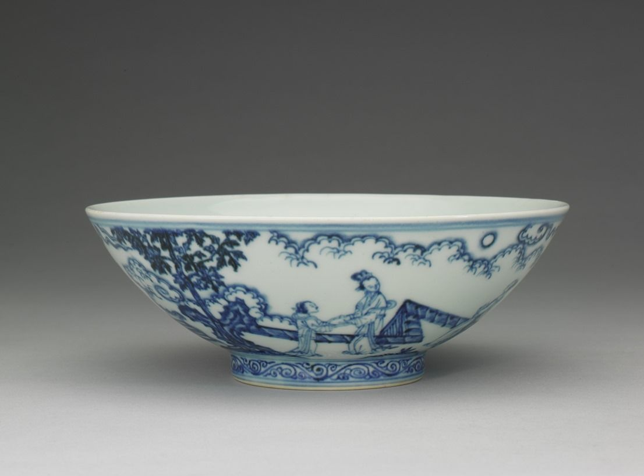}
\includegraphics[width=0.32\linewidth]{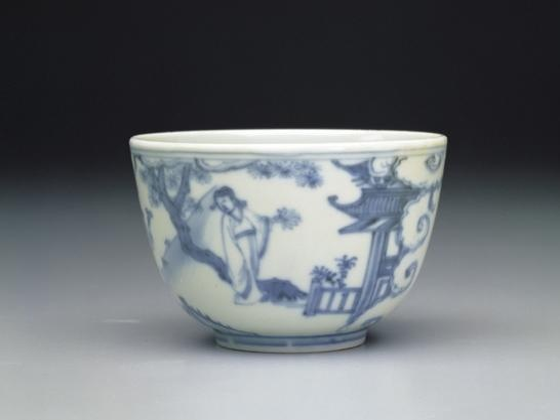}
\includegraphics[width=0.32\linewidth]{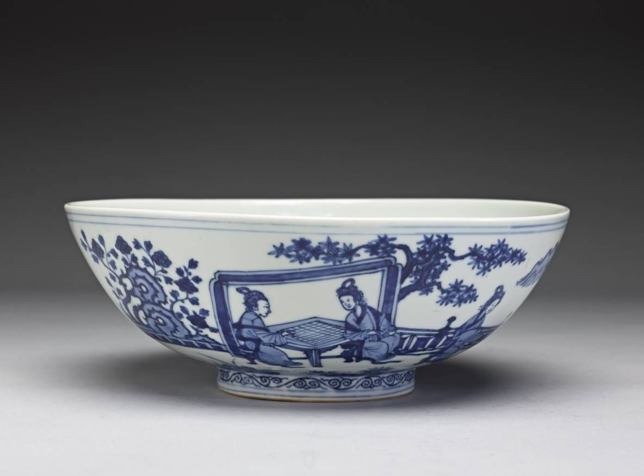}

Image search results:

  - Name: Ming Wanli Wucai Garlic-Mouth Vase with Dragon, Flowers and Birds
  
  - Dynasty: Ming
  
  - Reign: Wanli
  
  - Glaze Color: Wucai (Five-color)
  
  - Decoration: Dragon with flowers and birds
  
  - Form: Garlic-mouth vase
  
  - Description: The slightly contracted mouth forms a garlic shape, with a long neck, swollen belly, and shallow footring. The entire vessel is thick and heavy. It was made by forming separate sections and then joining them, therefore five seam lines appear on the mouth, neck, and belly. Both the inner footring and base are unglazed, with areas stained by dripped glaze. The clay body is coarse and rich in iron, causing unglazed parts to fire into an orange tone with iron speckles. The whole surface is covered with a white glaze; in addition to underglaze-blue linear decoration, further scrolls, entwined flowers, freehand birds, and angular patterns are applied in multicolored low-temperature enamels including red, yellow, orange, green, and black. The main motif is a dragon striding among flowers on the belly. Four dragons with large round eyes appear as if dancing and grabbing pearls, their strange and exaggerated shapes carrying a cartoon-like appeal. A six-character kaishu mark “Made in the Wanli period of the Great Ming Dynasty” is written in underglaze blue on the mouth rim. The use of many low-temperature overglaze colors began in the Xuande reign, and by the Jiajing and Wanli periods, potters applied various glazes onto complex patterns to create a lively and vibrant visual effect.
  
  - Source: National Palace Museum, Taipei
  
  - Similarity: 0.935499

---

Image search results:

  - Name: Ming Wanli Wucai Garlic-Mouth Vase with Dragon, Flowers and Birds
  
  - Dynasty: Ming
  
  - Reign: Wanli
  
  - Glaze Color: Wucai (Five-color)
  
  - Decoration: Dragon with flowers and birds
  
  - Form: Garlic-mouth vase
  
  - Description: Same as above.
  
  - Source: National Palace Museum, Taipei
  
  - Similarity: 0.93520796

---

Image search results:

  - Name: Qing Qianlong Famille-Rose Vase with Floral Phoenix Motif and Ruyi Handles
  
  - Dynasty: Qing
  
  - Reign: Qianlong
  
  - Glaze Color: Famille-rose
  
  - Decoration: Floral motifs
  
  - Form: Vase
  
  - Description: Celestial-globe form with flared mouth, long neck, rounded shoulders, and swollen belly, with a pair of ruyi-shaped handles on the sides of the neck. Flat base, outward-turned footring, and gold-painted rim. The body is decorated in famille-rose enamels over a white ground. Below the rim is a pomegranate motif; the neck shows lotus, bats, and ancient bells linked together, interspersed with floral scrolls. Two gilt string bands encircle the shoulder, framing multicolored patterns and leading to a ruyi-cloud border. The ruyi handles are linked with wan symbols and pendant ribbons attached to the shoulders. The belly features paired phoenixes between musical stones and lotus flowers, surrounded by dense floral scrolls. Near the bottom is a stylized lotus-petal band, above a ruyi-cloud border and a string line encircling the foot. A square six-character seal mark “Made in the Qianlong period of the Great Qing Dynasty” is written in underglaze blue.
  - Source: National Palace Museum, Taipei
  
  - Similarity: 0.9320861

\textbf{Assistant:} According to the results returned by the tool, the shape, decorative motifs, glaze color, and reign mark of this blue-and-white figure-decorated flared-rim bowl all correspond to typical characteristics of the Kangxi period of the Qing dynasty. It can be clearly distinguished from porcelain produced during the Xuande, Jiajing, and Wanli periods of the Ming dynasty, particularly in terms of the hue of the cobalt blue, the refinement of the painting, and the proportional maturity of the vessel form. These features reflect the highly advanced craftsmanship of Kangxi porcelain. Based on the narrative figure motif and the overall vessel shape, it can be confirmed as a work from the Kangxi imperial kilns.

Next, I intend to invoke the search\_text tool and input a detailed description of this bowl to further compare it with typical features of Kangxi blue-and-white porcelain, in order to verify its period attribution and standardized name.
\begin{lstlisting}[style=casestyle]
<tool_call>
{"name": "search_text", "arguments": {"query": "Qing Kangxi blue-and-white figure-decorated flared-rim bowl"}}
</tool_call>
\end{lstlisting}
\textbf{User:} Successfully found the following content:

\includegraphics[width=0.3\linewidth]{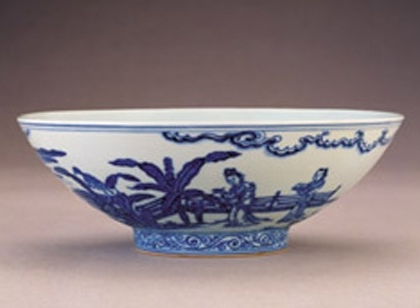}

Text search results:

 - Text: Foot diameter 2.7 cm. The center of the bowl is decorated with a Taiji (yin–yang) motif. The depiction of the yin–yang fishes is similar to the piece mentioned above, but the painting is less refined. The outer ring’s clockwise thin line is relatively smooth. The cobalt-blue color is rich. Within the double circles at the base center is an unrecognizable grass-style mark.
 
 - Title: Fragment of a green-glazed exterior bowl with blue-and-white interior, Qing Kangxi
 
 - Source: Authentication of Ming and Qing Porcelain Motifs — Decorative Motifs Volume
 
 - Similarity: 0.8

---

Text search results:

 - Text: Flared rim, deep belly, and footring. A band of floral motifs is painted along the inner rim. Inside the double blue rings at the center is a painting of rocks, flowers, and birds. The exterior wall depicts egrets in a lotus pond, with luxuriant lotuses rendered in clear layers. The lotus leaves show a sense of front-and-back yin–yang shading. The fine lines outline flowing water gracefully. Nine egrets, each in a different pose, are accurately drawn; light cobalt washes outside the outlines enhance expression with an elegant touch. The body is thin and fine, the glaze glossy and evenly applied. Within the double blue circles at the base is the kaishu mark “Made in the Jiajing period of the Great Ming,” yet it is a refined work from Kangxi folk kilns.
 
 - Title: Flared-rim deep-belly bowl with footring
 
 - Source: Complete Collection of Porcelain Collected in Jiangxi — Qing Dynasty, Volume I
 
 - Similarity: 0.7787347755736039

---

Image search results:

 - Name: Blue-and-white bowl with “banana leaves and ladies composing poetry” and Xuande reign mark, Ming
 
 - Dynasty: Ming
 
 - Reign: Xuande
 
 - Glaze Color: Blue-and-white
 
 - Decoration: Ladies
 
 - Form: Bowl
 
 - Description: A band of scroll motifs is painted around the footring. Six cobalt-blue lines are added between the rim and the foot. The cobalt blue is heavy, with brownish-green speckles. The body is thick, fine, white, and compact, with iron spots of different sizes. The base bears a two-column six-character kaishu mark “Made in the Xuande period of the Great Ming,” enclosed by double circles. The theme of this bowl is “lady with banana leaves composing poetry.” Due to somewhat thick brushwork, the lady appears with bold eyebrows and a full face, lacking delicate elegance, yet the painting is still fine and remains a high-quality work.
 
 - Source: National Palace Museum, Taipei
 
 - Similarity: 0.19999999999999996
 
\textbf{Assistant:} Based on the analysis, this blue-and-white flared-rim bowl with figure decoration is a typical porcelain piece from the Kangxi period of the Qing dynasty. Its vessel form, decorative motifs, glaze color, and reign mark are all consistent with the characteristics of imperial-kiln wares from that period. In conclusion, it should be identified as “Qing Kangxi Blue-and-White Flared-Rim Bowl with Narrative Figure Decoration.”
\begin{lstlisting}[style=casestyle]
<tool_call>
{"name": "image_zoom_in_tool", "arguments": {"index": 1, "bbox_2d": [112, 114, 826, 781], "label": "narrative figure motif"}}
</tool_call>
\end{lstlisting}

\textbf{User:} Successfully zoomed in on the image to the region labeled “narrative figure motif” at [112, 114, 826, 781].

\includegraphics[width=0.5\linewidth]{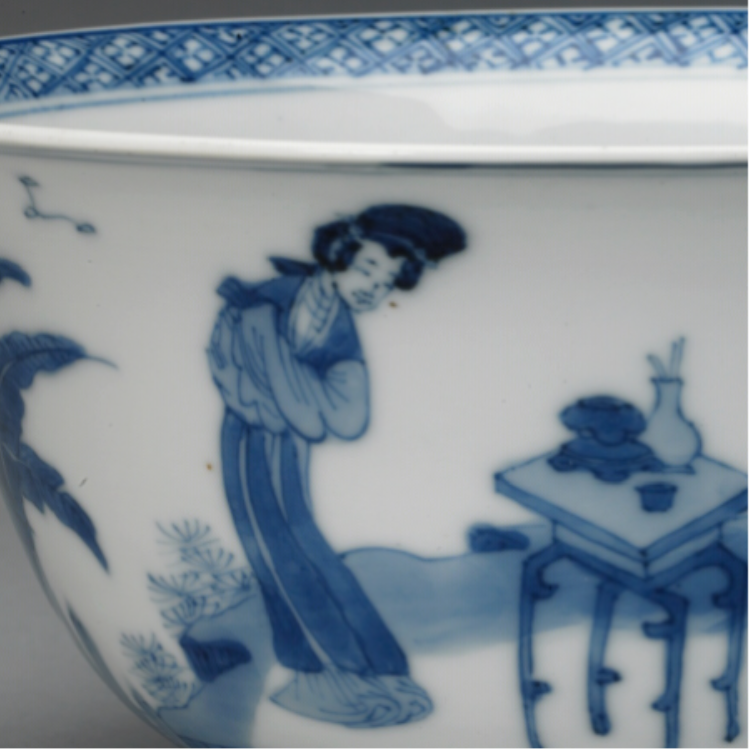}

\textbf{Assistant:} Based on the analysis, this porcelain piece is a typical work from the Kangxi period of the Qing dynasty. Its vessel form, decorative motifs, glaze color, and reign mark all conform to the characteristics of imperial-kiln wares from this period. It should therefore be identified as “Qing Kangxi Blue-and-White Flared-Rim Bowl with Figure Decoration.” 
\begin{lstlisting}[style=casestyle]
<answer>
Qing Kangxi Blue-and-White Flared-Rim Bowl with Figure Decoration
</answer>
\end{lstlisting}
\end{tcolorbox}

\begin{tcolorbox}[title=White-Glazed Standing Guanyin{,} Dehua Kiln{,} Ming Dynasty (\zhtext{明{ }德化窑白釉观音立像})]
\includegraphics[width=0.7\linewidth]{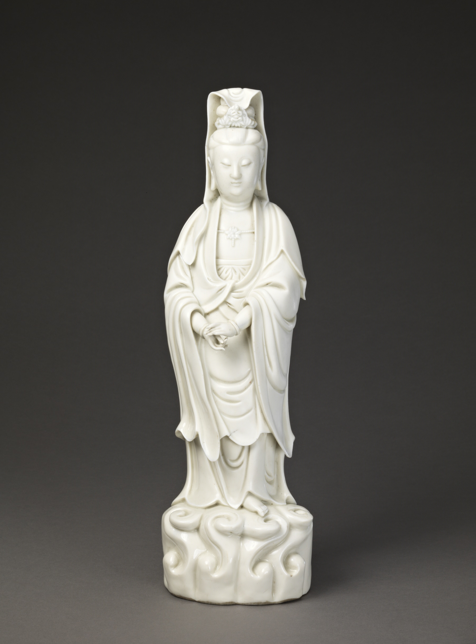}

\textbf{User:} Please provide a standardized name for this porcelain piece.

After your analysis, choose to either invoke a tool or respond directly, and place the final identification result within the \texttt{<answer></answer>} tags.

\textbf{Assistant:} This porcelain piece is a white-glazed standing Guanyin figure. Its glaze is lustrous and creamy, displaying the characteristic “ivory white” tone that is a defining feature of Dehua white porcelain. The sculptural workmanship is refined: the Guanyin has a dignified and compassionate expression, the drapery flows naturally, and the pedestal is carefully modeled, all of which correspond to typical features of Ming-dynasty Dehua kiln figures. Based on its style and craftsmanship, it can be inferred that this work was produced by the Dehua kilns during the Ming dynasty.

To further verify its kiln origin and chronological attribution, I intend to invoke the search\_image tool and input the image of this piece, in order to retrieve similar museum collections and assist in the identification.
\begin{lstlisting}[style=casestyle]
<tool_call>
{"name": "search_image", "arguments": {"index": 1}}
</tool_call>
\end{lstlisting}

\textbf{User:} Successfully found the following content: 

\includegraphics[width=0.32\linewidth]{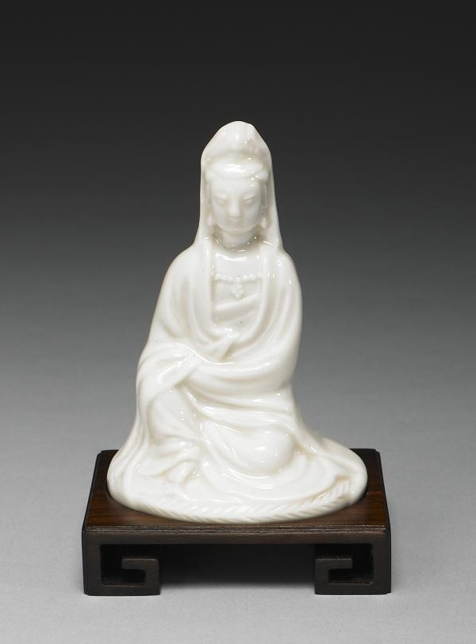}
\includegraphics[width=0.26\linewidth]{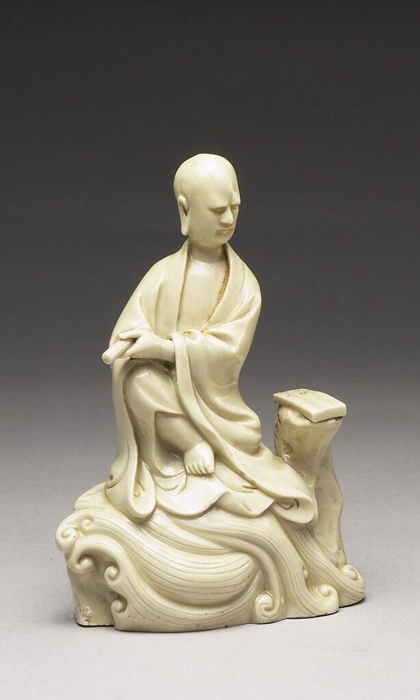}
\includegraphics[width=0.32\linewidth]{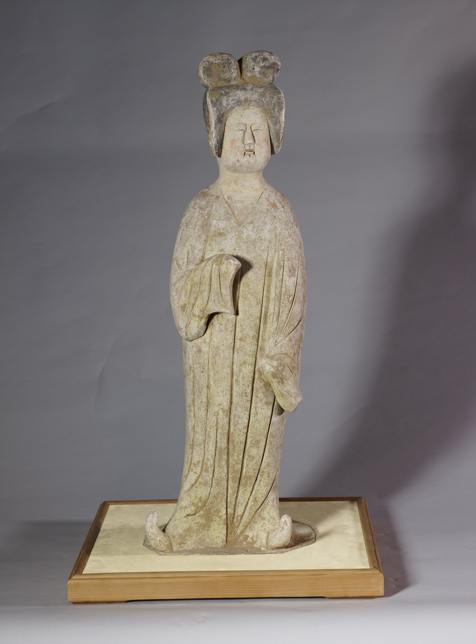}

Image search results:

 - Name: Qing Dehua Kiln Guanyin
 
 - Dynasty: Qing
 
 - Glaze Color: White glaze
 
 - Form: Guanyin figure
 
 - Description: Guanyin, “White-Robed Guanyin,” coated in white glaze, hollow inside. The face is oval like a goose egg, with delicate features. Wearing a white robe, a chest necklace, and a white hood, seated in a relaxed pose.
 
 - Source: National Palace Museum, Taipei
 
 - Similarity: 0.9279872

---

Image search results:

 - Name: 16th–18th Century White-Porcelain Arhat Figure
 
 - Glaze Color: White porcelain
 
 - Form: Arhat figure
 
 - Description: The glaze is ivory-white with a slightly yellow tint, with multiple pinholes across the surface. The figure is modeled seated by the water, with a bald head, long ears, and bare feet. Draped in a robe exposing the chest, the folds are fluently sculpted. The relaxed pose shows the left hand holding a sutra scroll and the right hand tucked within the sleeve. The face features pronounced brows and nose ridge, expressing contemplation. Accompanied by a wooden base.
 
 - Source: National Palace Museum, Taipei
 
 - Similarity: 0.8567522

---

Image search results:

 - Name: Tang Painted Grey-Terracotta Female Figure
 
 - Dynasty: Tang
 
 - Glaze Color: Painted
 
 - Decoration: Lady motif
 
 - Form: Tomb figure
 
 - Description: The painted garment on this grey terracotta female figure has mostly flaked away, yet the form still fully expresses the features of Tang court women—“large hair buns, loose robes,” and a plump figure. She stands upright with head raised, long eyes and small mouth, wearing a loose, pointed-collar robe draping from the neck. The lightly incised sleeve and fold lines resemble expressive ink strokes, accentuating the rounded body. The hem reaches the ground, with pointed shoes peeking out. The right arm lifts a wide sleeve to the chest, with the left hand dropping slightly, and the calm expression enhances the graceful bearing. The full face framed by a double-bun hairstyle embodies the Tang ideal of beauty. The era favored full-bodied women with confident, healthy elegance, a trend that continued into the Late Tang and Five Dynasties.
 
 - Source: National Palace Museum, Taipei
 
 - Similarity: 0.84899557
 
\textbf{Assistant:} Based on the analysis, the characteristics of this porcelain piece closely match the typical glaze tone, vessel form, and craftsmanship of Dehua ware. Further verification through tool results confirms that its style, glaze, and production techniques are consistent with those of Dehua kilns from the Ming dynasty.
\begin{lstlisting}[style=casestyle]
<tool_call>
{"name": "search_text", "arguments": {"query": "Ming dynasty Dehua kiln white-glazed standing Guanyin figure"}}</tool_call>
\end{lstlisting}
\textbf{User:}  Successfully found the following content: 

\includegraphics[width=0.32\linewidth]{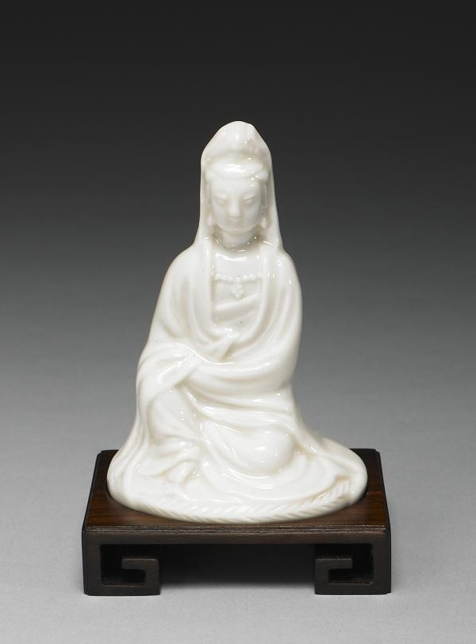}

Text search results:

 - Text: Ming, 17th century, Dehua white-glazed seated Guanyin figure, showing a Guanyin statue with hands hidden within the folds of her robe.
 
 - Title: Ming, 17th Century Dehua White-Glazed Seated Guanyin Figure
 
 - Source: Christie’s 2015 Spring Auction Catalogue — Important Chinese Ceramics
 
 - Similarity: 0.8

---

Text search results:

 - Text: White porcelain from the Dehua kilns of the Ming dynasty is distinctive, especially the religious sculptures of Guanyin, Bodhidharma, and other figures created by the master sculptor He Chaozong.
 
 - Title: White Porcelain of the Ming Dynasty Dehua Kilns
 
 - Source: Palace Museum Ceramic Encyclopedia
 
 - Similarity: 0.6398200054682729

---

Image search results:

 - Name: Qing Dehua Kiln Guanyin
 
 - Dynasty: Qing
 
 - Glaze Color: White glaze
 
 - Form: Guanyin figure
 
 - Description: Guanyin, “White-Robed Guanyin,” coated in white glaze, hollow inside. The face is oval like a goose egg, with delicate features. Wearing a white robe, a chest necklace, and a white hood, seated in a relaxed pose.
 
 - Source: National Palace Museum, Taipei
 
 - Similarity: 0.19999999999999996

\textbf{Assistant:} Based on the information returned by the tool, this porcelain piece is a white-glazed standing Guanyin figure from the Dehua kilns of the Ming dynasty. Its glaze, form, and workmanship are consistent with the typical characteristics of this period. It can therefore be identified as “Ming Dehua Kiln White-Glazed Standing Guanyin Figure.” I will now further zoom in on the details for confirmation.
\begin{lstlisting}[style=casestyle]
<tool_call>
{"name": "image_zoom_in_tool", "arguments": {"index": 1, "bbox_2d": [112, 114, 670, 1004], "label": "Guanyin"}}
</tool_call>
\end{lstlisting}

\textbf{User:} Successfully zoomed in on the region labeled “Guanyin” at [112, 114, 670, 1004].

\includegraphics[width=0.5\linewidth]{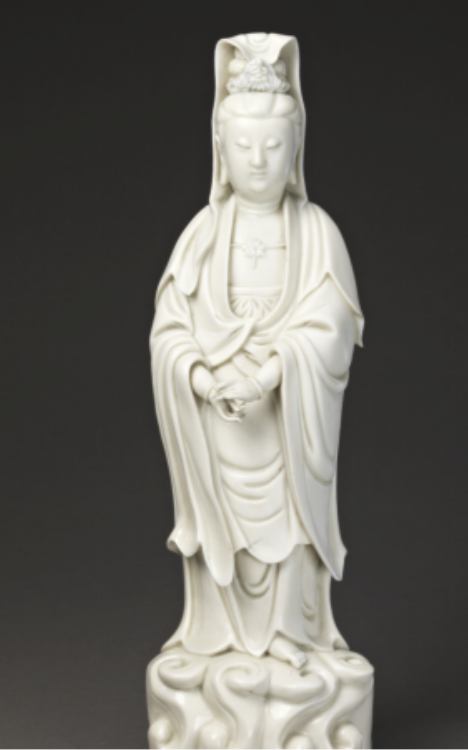}

\textbf{Assistant:} Based on the analysis of the porcelain and the verification through the tool results, this piece is a white-glazed standing Guanyin figure produced by the Dehua kilns in the Ming dynasty. It exhibits the typical glaze tone, vessel form, and craftsmanship associated with that period, and thus corresponds to the standard designation “Ming Dehua Kiln White-Glazed Standing Guanyin Figure.”
\begin{lstlisting}[style=casestyle]
<answer>
Ming Dehua Kiln White-Glazed Standing Guanyin Figure
</answer>
\end{lstlisting}
\end{tcolorbox}

\begin{tcolorbox}[title=Tendrilled Lotus Patternp (\zhtext{缠枝莲纹})]
\includegraphics[width=0.48\linewidth]{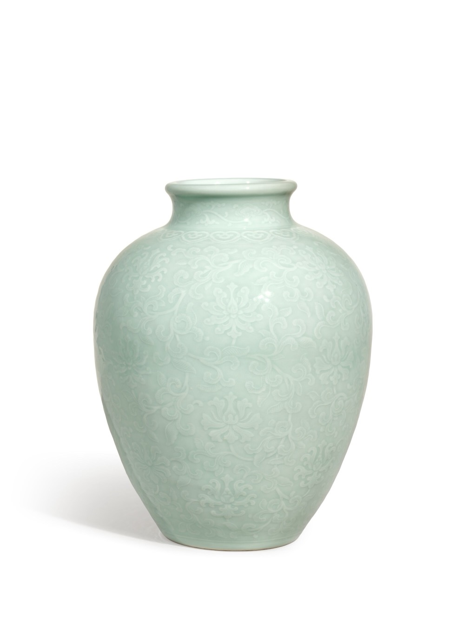}
\includegraphics[width=0.48\linewidth]{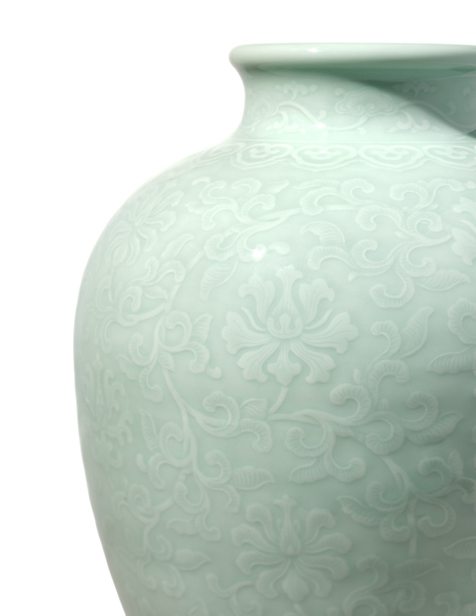}

\textbf{User:} Please determine the decorative motif of this antique.

After your analysis, choose to either invoke a tool or respond directly, and place the final identification result within the \texttt{<answer></answer>} tags.

\textbf{Assistant:} The decorative motif of this porcelain piece is a tendrilled lotus pattern. Its characteristics include gracefully flowing vine scrolls and full, symmetrical lotus blossoms, creating a complex yet orderly overall composition. This style is typical of imperial ware from the Yongzheng period of the Qing dynasty. From the image, the lines of the motif appear delicate and precisely rendered, demonstrating a high level of carved-decoration craftsmanship.

To further verify its stylistic period and craftsmanship, I intend to invoke the search\_image tool and input the index of this porcelain image (starting from 1) in order to search for similar examples in ancient collections that can assist in identification.
\begin{lstlisting}[style=casestyle]
<tool_call>
{"name": "search_image", "arguments": {"index": 1}}
</tool_call>
\end{lstlisting}

\textbf{User:} Successfully found the following content: 

\includegraphics[width=0.32\linewidth]{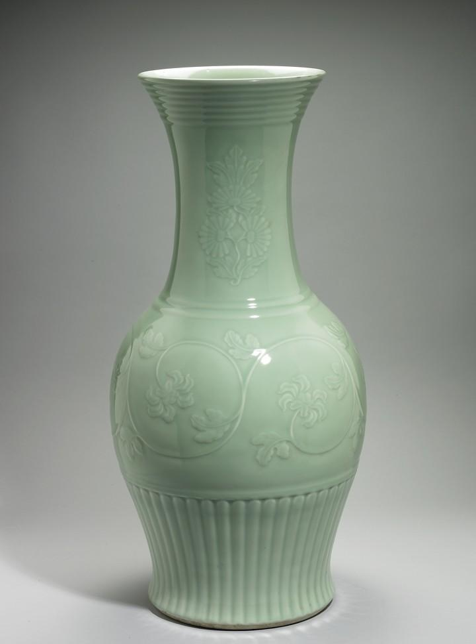}
\includegraphics[width=0.32\linewidth]{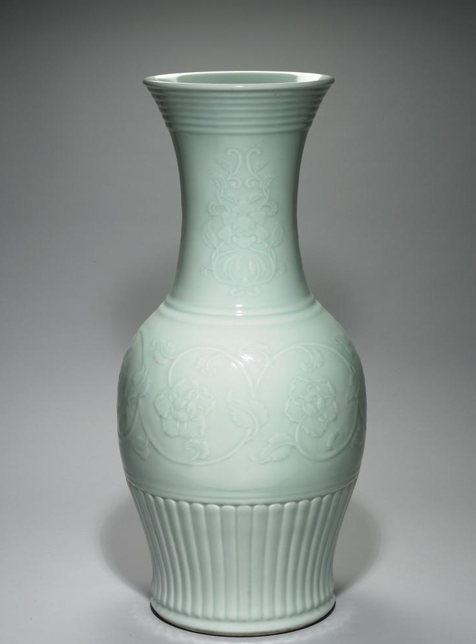}
\includegraphics[width=0.32\linewidth]{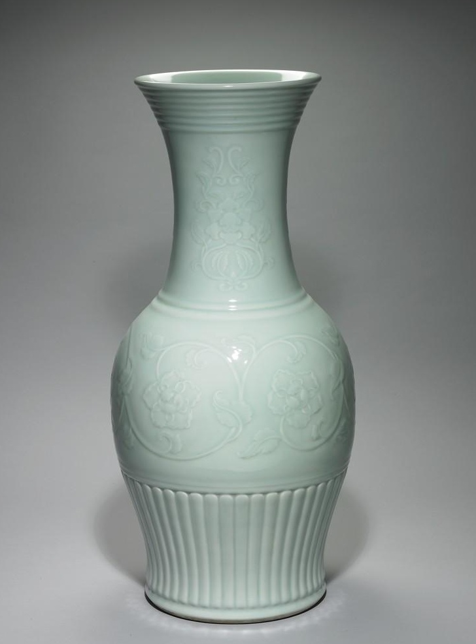}

Image search results:

 - Name: Qing Yongzheng Imitation Longquan Kiln Phoenix-Tail Vase
 
 - Dynasty: Qing
 
 - Reign: Yongzheng
 
 - Glaze Color: Celadon glaze
 
 - Decoration: Tendrilled floral motif
 
 - Form: Vase
 
 - Description: Flared mouth with straight neck, swollen body tapering downward and flaring outward at the foot. The entire vessel is coated in a celadon glaze. Several string lines encircle the mouth rim. The neck and belly are each decorated with floral blossoms and tendrilled scroll motifs, with a band of chrysanthemum-petal pattern around the lower belly. The glaze tone is pale and evenly applied. Inside the footring is a six-character, three-column seal-script mark: “Made in the Yongzheng period of the Great Qing.”
 
 - Source: National Palace Museum, Taipei
 
 - Similarity: 0.9185529

---

Image search results:

 - Name: Qing Yongzheng Winter-Green Celadon Vase with Incised Decoration
 
 - Dynasty: Qing
 
 - Reign: Yongzheng
 
 - Glaze Color: Winter-green glaze
 
 - Decoration: Incised motif
 
 - Form: Vase
 
 - Description: Flared mouth with straight neck, swollen body tapering downward and flaring outward at the foot. The entire vessel is coated in a celadon glaze. Several string lines encircle the mouth rim. The neck and belly are each decorated with floral blossoms and tendrilled scroll motifs, with a band of chrysanthemum-petal pattern around the lower belly. The glaze tone is pale and evenly applied. Inside the footring is a six-character, three-column seal-script mark: “Made in the Yongzheng period of the Great Qing.”
 
 - Source: National Palace Museum, Taipei
 
 - Similarity: 0.9150853

---

Image search results:

 - Name: Qing Yongzheng Winter-Green Celadon Vase with Incised Decoration
 
 - Dynasty: Qing
 
 - Reign: Yongzheng
 
 - Glaze Color: Winter-green glaze
 
 - Decoration: Incised motif
 
 - Form: Vase
 
 - Description: Flared mouth with straight neck, swollen body tapering downward and flaring outward at the foot. The entire vessel is coated in a celadon glaze. Several string lines encircle the mouth rim. The neck and belly are each decorated with floral blossoms and tendrilled scroll motifs, with a band of chrysanthemum-petal pattern around the lower belly. The glaze tone is pale and evenly applied. Inside the footring is a six-character, three-column seal-script mark: “Made in the Yongzheng period of the Great Qing.”
 
 - Source: National Palace Museum, Taipei
 
 - Similarity: 0.9142115
 
\textbf{Assistant:} Based on the results returned by the tool, the decorative motif on this porcelain piece is indeed a tendrilled lotus pattern. Its stylistic features closely correspond to those found on imperial wares from the Yongzheng period of the Qing dynasty, further substantiating the accuracy of its period attribution and craftsmanship.
\begin{lstlisting}[style=casestyle]
{"name": "image_zoom_in_tool", "arguments": {"index": 2, "bbox_2d": [112, 114, 888, 987], "label": "tendrilled lotus pattern"}}
\end{lstlisting}

\textbf{User:} Successfully zoomed in on the region labeled “tendrilled lotus pattern” at [112, 114, 888, 987].

\includegraphics[width=0.6\linewidth]{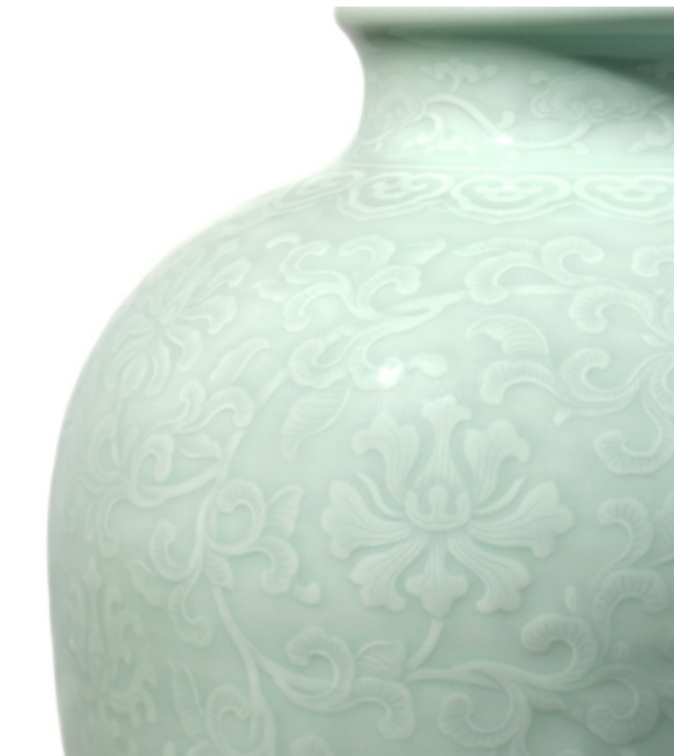}

\textbf{Assistant:} The decorative motif of this porcelain piece is a typical tendrilled lotus pattern. The scrolling lines are graceful and fluent, and the lotus blossoms are rendered with delicate precision, fully reflecting the refined craftsmanship and aesthetic pursuit of imperial porcelain from the Yongzheng period of the Qing dynasty.
\begin{lstlisting}[style=casestyle]
<answer>
Tendrilled lotus pattern
</answer>
\end{lstlisting}
\end{tcolorbox}

\section{Prompts} \label{supp_sec:prompt}
\subsection{Prompt for Metadata Enrichment}
\begin{lstlisting}[style=promptstyle, caption={Prompt for generating multiple-choice options}]
Please describe the Chinese porcelain shown in the image: <image>. 
Then revise the description according to the human experts annotation:<description>. 
The final description should comprehensively cover six attributes: dynasty, reign period, kiln site, glaze color, decorative motif, and vessel shape.
\end{lstlisting}

\subsection{Prompt for VQA Data Generation}
% Given the expert-enhanced description of a porcelain artifact:

% <full_description>

% Your task is to extract four categories of information from this description: 
% (1) Dynasty, (2) Vessel Shape, (3) Glaze Color, and (4) Decorative Motifs.

% Construct one QA pair at a time.  
% For the current step, answer **only one** of the four categories.  

% Format:
% Q: A question that asks specifically about one category (e.g., dynasty, vessel shape, glaze color, or decorative motifs).
% A: A concise but complete answer extracted from the description, including all relevant descriptive details.

% Do not include information from the other three categories in the answer.  
% Do not infer beyond the given description; extract only what is stated.

% Generate the QA pair for the following category:
% <category>
\begin{lstlisting}[style=promptstyle, caption={Prompt for generating multiple-choice options}]
Given the expert-enhanced description of a porcelain artifact:

<full_description>

Your task is to extract four categories of information from this description: 
(1) Dynasty, (2) Vessel Shape, (3) Glaze Color, and (4) Decorative Motifs.

Construct one QA pair at a time.  
For the current step, answer **only one** of the four categories.  

Format:
Q: A question that asks specifically about one category (e.g., dynasty, vessel shape, glaze color, or decorative motifs).
A: A concise but complete answer extracted from the description, including all relevant descriptive details.

Do not include information from the other three categories in the answer.  
Do not infer beyond the given description; extract only what is stated.

Generate the QA pair for the following category:
<category>
\end{lstlisting}

\subsection{Prompt for Generating Multiple-choice Options} \label{supp_subsec:prompt_options}

\begin{lstlisting}[style=promptstyle, caption={Prompt for generating multiple-choice options}]
<image>You are an expert in porcelain-related question design. Below is a porcelain-related question and its correct answer. Please transform this question into a multiple-choice question with four options, where one option is correct and the other three are plausible but incorrect.

Please output strictly in the following format:
<A>Content of option A</A>
<B>Content of option B</B>
<C>Content of option C</C>
<D>Content of option D</D>
<answer>Letter of the correct option</answer>

Notes:
- The options must be relevant to the given question and answer.
- The incorrect options should be misleading but distinguishable from the correct one.
- Do not add any other text or explanations outside the specified format.
\end{lstlisting}

\subsection{Prompt for LLM-as-a-Judge in Evaluation} \label{supp_subsec:prompt_llm-as-a-judge_eval}
\begin{lstlisting}[style=promptstyle, caption={Prompt for LLM-as-a-Judge in Evaluation}]
You are an expert reviewer in the field of antique Chinese porcelain, specialized in evaluating the accuracy of model identification results.
Please compare the following "reference answer" and "model output" and produce a comprehensive evaluation according to the rules below across 6 dimensions.

---

### Scoring requirements:
For each of the seven dimensions listed below, assign an individual score between 0 and 1. Adopt a conservative scoring style.

### Detailed scoring rules:
1.**Dynasty accuracy**
Whether the output mentions and correctly states the dynasty to which the porcelain belongs, and whether the provided dynasty information is sufficiently precise.
2.**Reign period accuracy**
Whether the output mentions and correctly states the imperial reign (e.g., Kangxi, Qianlong); note any deviation or omission.
3.**Kiln site accuracy**
Whether the output mentions and correctly states the kiln site characteristics (e.g., Jingdezhen, Ru Kiln); note any deviation or omission.
4.**Glaze color accuracy**
Whether the output mentions and correctly states the glaze/color characteristics (e.g., blue-and-white, famille-rose, red-ground with green enamels); note any deviation or omission.
5.**Decoration / Motif accuracy**
Whether the output correctly describes the decorative motifs or subjects (e.g., dragons and phoenixes, floral patterns, human figures, cloud patterns), and whether the motif matches the stylistic expectations of the claimed period.
6.**Form / Vessel-type accuracy**
Whether the output reasonably identifies and describes the vessel form, and whether this conforms with the reference answer.

Note: If a particular dimension is absent in the reference answer, mark it with -1 to indicate missing data.

Please first provide the reasoning for each score, and then place the final numeric scores inside the following tags in order:
<Dynasty>...</Dynasty>
<Reign>...</Reign>
<Kiln>...</Kiln>
<Color>...</Color>
<Motif>...</Motif>
<Shape>...</Shape>

For example:
<Dynasty>1.0</Dynasty>
<Reign>0.6</Reign>
<Kiln>-1.0</Kiln>
<Color>1.0</Color>
<Motif>0.0</Motif>
<Shape>0.8</Shape>

Reference answer: {ground_truth}
Model output: {prediction}
\end{lstlisting}

\subsection{Prompt for LLM-as-a-Judge in Training} \label{supp_subsec:prompt_llm-as-a-judge_train}
The training-time LLM-as-a-Judge prompt is largely aligned with the evaluation-time version; the only difference is the addition of a \texttt{Format consistency} criterion to mitigate potential reward hacking during reinforcement learning.
\begin{lstlisting}[style=promptstyle, caption={Prompt for LLM-as-a-Judge in Training}]
You are an expert reviewer in the field of antique Chinese porcelain, specialized in evaluating the accuracy of model identification results.
Please compare the following "reference answer" and "model output" and produce a comprehensive evaluation according to the rules below across seven dimensions.

---

### Scoring requirements:
For each of the seven dimensions listed below, assign an individual score between 0 and 1. Adopt a conservative scoring style.

### Detailed scoring rules:
1.**Format consistency**
Whether the output strictly follows the naming order: "Dynasty", "Reign period", "Kiln site", "Glaze color", "Decoration motif", "Vessel shape". Some elements may be missing, but the order must not be disrupted. The model output must contain only a single standard name; it must not include any explanatory, descriptive, or other additional text. If the order is correct but some fields are missing, 1 score may still be given.
2.**Dynasty accuracy**
Whether the output mentions and correctly states the dynasty to which the porcelain belongs, and whether the provided dynasty information is sufficiently precise.
3.**Reign period accuracy**
Whether the output mentions and correctly states the imperial reign (e.g., Kangxi, Qianlong); note any deviation or omission.
4.**Kiln site accuracy**
Whether the output mentions and correctly states the kiln site characteristics (e.g., Jingdezhen, Ru Kiln); note any deviation or omission.
5.**Glaze color accuracy**
Whether the output mentions and correctly states the glaze/color characteristics (e.g., blue-and-white, famille-rose, red-ground with green enamels); note any deviation or omission.
6.**Decoration / Motif accuracy**
Whether the output correctly describes the decorative motifs or subjects (e.g., dragons and phoenixes, floral patterns, human figures, cloud patterns), and whether the motif matches the stylistic expectations of the claimed period.
7.**Form / Vessel-type accuracy**
Whether the output reasonably identifies and describes the vessel form, and whether this conforms with the reference answer.

Note: If a particular dimension is absent in the reference answer, mark it with -1 to indicate missing data.

Please first provide the reasoning for each score, and then place the final numeric scores inside the following tags in order:
<Format>...</Format>
<Dynasty>...</Dynasty>
<Reign>...</Reign>
<Kiln>...</Kiln>
<Color>...</Color>
<Motif>...</Motif>
<Shape>...</Shape>

For example:
<Format>1.0</Format>
<Dynasty>1.0</Dynasty>
<Reign>0.6</Reign>
<Kiln>-1.0</Kiln>
<Color>1.0</Color>
<Motif>0.0</Motif>
<Shape>0.8</Shape>

Reference answer: {ground_truth}
Model output: {prediction}
\end{lstlisting}

\section{Detailed Configurations} \label{supp_sec:detailed_config}
 
\noindent\textbf{SFT Setup.}
We conducted SFT using the \textit{ms-swift}~\cite{ms_swift} framework for $1$ epoch with a learning rate of $1\times10^{-5}$, a batch size of $16$, and gradient accumulation over $4$ steps. We used the default AdamW optimizer and set the maximum sequence length to $8192$.

\noindent\textbf{RL Setup.} During the RL training, we use the AdamW with a learning rate of $1\times10^{-6}$, a batch size of 128 with 16 rollouts for each prompt, and a KL coefficient of $\beta=0$. 

\noindent\textbf{Tool Setup.} The size of the initial image sent to the agent is set to be maximum $313{,}600$ pixels. For the image zoom-in tool, the zoom-in operation is performed on the initial image by mapping the \texttt{bbox} back to the original image size.
We set $k=3$ for both search image and search text, and the coefficient $\alpha$ is set as $0.2$. The maximum number of tool invocations per query is set to $4$ to balance between reasoning depth and computational efficiency. 

\section{Comparison with Existing Datasets}\label{supp_sec:comp_data}
% 下面我们给出我们提出的CiQi-VQA dataset和已有的一些瓷器相关数据集的比较，见表2.
Below, we present a comparison between our proposed CiQi-VQA dataset and several existing porcelain-related datasets, as shown in Table~\ref{tab:dataset_comp}. We report the number of porcelain specimens included in each dataset, as well as the number of categories covered under six attributes. A “---” indicates that the dataset does not provide statistics for that attribute. As shown in the comparison, our dataset contains the largest scale and offers the most fine-grained attribute coverage among all datasets.
\begin{table*}[htbp]
\centering
\caption{Dataset Comparison.}
\label{tab:dataset_comp}
\begin{tabular}{lccccccc}
\noalign{\hrule height 0.8pt}
Datasets & Porcelain Specimens  & Dynasties & Reigns & Kilns & Colors & Motifs &Shapes \\
\hline
Ling et al.~\cite{Ling2025MultiTask} & 5,993 & 2 & --- & 10 & 8 & --- & 12 \\
Ma et al.~\cite{9492261} & 5,624 & --- & --- & --- & --- & --- & 7 \\
Sun et al.~\cite{SUN202314246} & 232 & 3 & --- & --- & --- & --- & 1 \\
Yang et al.~\cite{9898614} & 2,750 & --- & --- & --- & 7 & --- & --- \\
Hu et al.~\cite{hu2025integrating} & 8,213 & 6 & --- & --- & 20 & 6 & 7 \\
\midrule
CiQi-VQA (Ours) & 29,596 & 38 & 42 & 43 & 246 & 248 & 158 \\
\noalign{\hrule height 0.8pt}
\end{tabular}
\end{table*}

\begin{table*}[ht]
\centering
\caption{Comparative experiments on the Hu et al.~\cite{hu2025integrating} dataset.}
\label{tab:supp_experiments}
\begin{tabular}{lccc}
\noalign{\hrule height 0.8pt}
Model & Glaze/Kiln & Shape & Weighted Avg.$^\dagger$ \\
\hline
GPT-5~\cite{openai2025gpt5} & \underline{86.9} & \underline{91.4} & \underline{88.1} \\
GPT-4.1~\cite{openai2025gpt41} & 86.4 & 88.6 & 87.0 \\
GPT-4o~\cite{openai2024gpt4o} & 86.4 & \underline{91.4} & 87.7 \\
o3~\cite{openai2025introducingo3} & \underline{86.9} & \underline{91.4} & \underline{88.1} \\
Claude Opus 4~\cite{anthropic_claude4_systemcard_2025} & 84.4 & 90.0 & 85.9 \\
\midrule
Qwen2.5-VL-72B-Instruct~\cite{qwen25vl} & 82.4 & 88.6 & 84.0 \\
GLM-4.5V (106B)~\cite{glm45v_2025} & 85.4 & \underline{91.4} & 87.0 \\
InternVL3.5-241B-A28B-Flash~\cite{internvl35_2025} & 74.4 & 90.0 & 78.4 \\
Kimi-VL-A3B-Instruct (16B)~\cite{kimivl_2025} & 74.9 & 85.7 & 77.7 \\
Hu et al.~\cite{hu2025integrating} & --- & --- & 70.0 \\
\midrule
CiQi-Agent (Ours, 7B) & \textbf{87.4} & \textbf{92.9} & \textbf{88.9} \\
\noalign{\hrule height 0.8pt}
\end{tabular}

\footnotesize{$^\dagger$Weighted by item counts: 
$w_\text{G}=199/269$, $w_\text{S}=70/269$.}
\end{table*}

\section{Additional Comparative Experiments}\label{supp_sec:add_exp}
% 我们在Hu et al.\cite{hu2025integrating}开源的数据集上进行了额外的对比试验，Hu et al.\cite{hu2025integrating}文中对于瓷器进行的是目标检测与识别任务，该文将瓷器的识别任务分成了器型、釉色/窑口（该文将釉色和窑口视为一类）与纹饰内容三类，每个大类又包含多个子类（详见表2）。在对比实验中，由于其开源的数据集不全，测试集没有标注，我们在其验证集上做了对比实验，并且由于其开源的数据集中，纹饰内容一项用的并不是瓷器图片，因此我们只在釉色/窑口与器型两个维度进行了对比实验。我们使用开源数据集，在这两个维度上生成了选择题，选项为该维度上所有的子类（釉色/窑口20类；器型6类），我们使用CiQi-Agent还有gpt系列，claude等9个MLLMs进行了对比试验，实验结果见表4。

We conducted additional comparative experiments using the open-source dataset provided by Hu et al.~\cite{hu2025integrating}. In their work, Hu et al.~\cite{hu2025integrating} performed object detection and recognition tasks on Chinese porcelains, defining three attributes for recognition: vessel shape, glaze color/kiln site (treated as a single attribute in their paper), and decorative motif. Each major attribute contains multiple sub-attributes (see Table~\ref{tab:dataset_comp} for details).

In our comparative study, because the released dataset is incomplete and the test set lacks annotations, we conducted the comparison using their validation set. We consider this comparison still meaningful because the paper states that both the validation set and the test set were obtained by randomly sampling $20$\% and $10$\% of the full dataset, respectively. Therefore, the results reported on the test set in the original paper are also informative for the validation set and should not deviate significantly. Additionally, in the open-source dataset, the decorative motif attribute is not represented by images of Chinese porcelains; therefore, we restricted our experiments to the glaze color/kiln site and vessel shape attributes.

We generated multiple-choice questions for these two attributes, with the options corresponding to all sub-attributes under each attribute ($20$ for glaze color/kiln site and $7$ for vessel shape). Specifically, the glaze color/kiln site attribute included $199$ questions, while the vessel shape attribute contained $70$ questions. We evaluated CiQi-Agent alongside nine MLLMs---including the GPT series, Claude, and others. The experimental results are summarized in Table~\ref{tab:supp_experiments}.

The experimental results show that on the open-source dataset provided by Hu et al.~\cite{hu2025integrating}, our CiQi-Agent consistently outperforms all baseline models. Moreover, its fine-grained classification capability significantly surpasses the approach described in Hu et al.~\cite{hu2025integrating}, achieving an average accuracy improvement of $18.9$\%. The small performance gap between our model and the other MLLMs is largely due to the simplicity of this dataset, which includes only $7$ vessel-shape categories and $20$ glaze categories. In contrast, our constructed dataset includes $100$+ of categories for both vessel shapes and glazes, making the task substantially more challenging.

% % \section*{References for Supplementary Material}
% \begingroup
%     \renewcommand{\refname}{References for Supplementary Material}
%     \renewcommand*{\backref}[1]{}%
%     \renewcommand*{\backrefalt}[4]{}%
%     \begin{thebibliography}{99}
%     \bibitem[S1]{S1Ma2021}
%     Jinjing Ma, Yongkang Peng, Wenfang Cheng, Meikang Qiu, and Yu Nie.
%     \newblock Identification method of ancient ceramics revision.
%     \newblock In \emph{2021 8th IEEE International Conference on Cyber Security and
%       Cloud Computing (CSCloud)/2021 7th IEEE International Conference on Edge
%       Computing and Scalable Cloud (EdgeCom)}, pages 213--218, 2021.
    
%     \bibitem[S2]{S2Sun2023}
%     Jingmin Sun, Hailiang Lu, Lei Qiao, Xiaolong Li, Kaihan Chen, and Weiran Cao.
%     \newblock Identification of porcelain ewers in tang, song, and yuan dynasties
%       by digital shape characterization.
%     \newblock \emph{Ceramics International}, 49(9, Part A):14246--14254, 2023.
    
%     \bibitem[S3]{S3Yang2022}
%     Yang Yang, Hui Wu, Dingguo Yu, and Chen Yang.
%     \newblock Ceramic type recognition algorithm based on ontology modeling and
%       transfer learning.
%     \newblock In \emph{2022 International Conference on Culture-Oriented Science
%       and Technology (CoST)}, pages 6--10, 2022.
    
%     \end{thebibliography}
% \endgroup

\end{document}